\PassOptionsToPackage{dvipsnames}{xcolor}
\documentclass{article}

\usepackage{microtype}
\usepackage{graphicx}
\usepackage{subfigure}
\usepackage{booktabs} %
\usepackage{float}
\usepackage{hyperref}

\usepackage[accepted]{icml2025}

\usepackage{amsmath}
\usepackage{amssymb}
\usepackage{mathtools}
\usepackage{amsthm}

\usepackage[capitalize,noabbrev]{cleveref}

\theoremstyle{plain}

\theoremstyle{definition}

\theoremstyle{remark}

\usepackage[textsize=tiny]{todonotes}

\usepackage[nolist]{acronym}
\usepackage{duckuments}
\usepackage{xspace}         %
\usepackage{tcolorbox}  %
\tcbuselibrary{skins}
\tcbuselibrary{breakable} %
\usepackage[dvipsnames]{xcolor}     %
\usepackage{inconsolata}
\usepackage{listings}
\usepackage{dirtytalk}  %
\usepackage{fancyvrb}
\usepackage{placeins}  %
\usepackage{longtable, array, multirow}  %

\icmltitlerunning{Insights into a radiology-specialised multimodal large language model with sparse autoencoders}

\begin{document}
\begin{acronym}
    \acro{SAE}{sparse autoencoder}
    \acro{LLM}{large language model}
    \acro{AI}{artificial intelligence}
    \acro{VLM}{vision--language model}
    \acro{MLLM}{multimodal large language model}
    \acro{CXR}{chest X-ray}
\end{acronym}

\definecolor{customblue}{HTML}{aaccff} 

\newcommand{\mairatwo}{MAIRA-2\xspace}
\newcommand{\llava}{LLaVA\xspace}

\newcommand{\mimiccxr}{MIMIC-CXR\xspace}
\newcommand{\cxr}{chest X-ray\xspace}

\newcommand{\topk}{TopKSAE\xspace}
\newcommand{\batchtopk}{BatchTopKSAE\xspace}
\newcommand{\matsae}{Matryoshka-SAE\xspace}

\newcommand{\nlatents}{16,384\xspace}

\newcommand{\hf}{Hugging Face\xspace}
\newcommand{\fone}{\ensuremath{F_1}\xspace}

\newcommand{\feature}[1]{$f_{#1}$}
\newcommand{\featexpl}[1]{\textit{`#1'}}

\newcommand{\balpha}{\boldsymbol{\alpha}}

\makeatletter
\newcommand{\wip}[1]{{\color{violet}#1}\@latex@warning{WIP segment}}
\newcommand{\@coloredcomment}[2]{{\color{#1}[#2]}\@latex@warning{#2}}
\newcommand{\TODO}[1]{\@coloredcomment{red}{TODO: #1}}
\newcommand{\daniel}[1]{\@coloredcomment{orange}{Daniel: #1}}
\newcommand{\felix}[1]{\@coloredcomment{orange}{Felix: #1}}
\newcommand{\kenza}[1]{\@coloredcomment{blue}{Kenza: #1}}
\newcommand{\steph}[1]{\@coloredcomment{purple}{Steph: #1}}
\newcommand{\shruthi}[1]{\@coloredcomment{pink}{Shruthi: #1}}
\newcommand{\tbd}[1]{{\color{magenta}#1}\@latex@warning{TBD: #1}}
\makeatother

\newtcolorbox{promptbox}[1][LLM Prompt]{
    breakable,
    enhanced,
    colback=customblue!25,
    colframe=customblue!75,
    arc=1mm,
    title=#1,
    fonttitle=\ttfamily\bfseries,
    fontupper=\ttfamily\footnotesize,
    coltitle=black,
    attach boxed title to top left={yshift=-1.5mm, xshift=2mm},
    boxed title style={colback=customblue!75, colframe=customblue!75}
}

\twocolumn[
\icmltitle{Insights into a radiology-specialised multimodal large language model with sparse autoencoders}

\icmlsetsymbol{equal}{*}

\begin{icmlauthorlist}
\icmlauthor{Kenza Bouzid}{hf}
\icmlauthor{Shruthi Bannur}{hf}
\icmlauthor{Felix Meissen}{hf}
\icmlauthor{Daniel Coelho de Castro}{hf}
\icmlauthor{Anton Schwaighofer}{hf}
\icmlauthor{Javier Alvarez-Valle}{hf}
\icmlauthor{Stephanie L. Hyland}{hf}
\end{icmlauthorlist}

\icmlaffiliation{hf}{Microsoft Research, Health Futures, Cambridge, United Kingdom}

\icmlcorrespondingauthor{Kenza Bouzid}{kenza.bouzid@microsoft.com}
\icmlcorrespondingauthor{Stephanie Hyland}{stephanie.hyland@microsoft.com}

\icmlkeywords{Machine Learning, AIW, mechanistic interpretability, radiology}

\vskip 0.3in
]

\printAffiliationsAndNotice{}  %

\begin{abstract}
Interpretability can improve the safety, transparency and trust of \ac{AI} models, which is especially important in healthcare applications where decisions often carry significant consequences.
Mechanistic interpretability, particularly through the use of \acp{SAE}, offers a promising approach for uncovering human-interpretable features within large transformer-based models. In this study, we apply \matsae to the radiology-specialised \acl{MLLM}, \mairatwo, to interpret its internal representations.
Using large-scale automated interpretability of the SAE features, we identify a range of clinically relevant concepts---including medical devices (e.g., line and tube placements, pacemaker presence), pathologies such as pleural effusion and cardiomegaly, longitudinal changes and textual features.
We further examine the influence of these features on model behaviour through steering, demonstrating directional control over generations with mixed success. 
Our results reveal practical and methodological challenges, yet they offer
initial insights into the internal concepts learned by \mairatwo---marking a step toward deeper mechanistic understanding and interpretability of a radiology-adapted \acl{MLLM}, and paving the way for improved model transparency.

We release the trained \acp{SAE} and interpretations: \url{https://huggingface.co/microsoft/maira-2-sae}.

\end{abstract}

\section{Introduction}
\label{sec:introduction}
Recent advancements in automated draft radiology reporting
\citep{bannur_maira-2_2024,zhou2024generalist,yang2024advancing,tu2024medpalmm,chen2024chexagent,hyland2023maira, wang2023metransformer,li2023dynamic} raise the potential of \ac{AI} systems to reduce radiologist workloads and improve operational efficiency~\citep{yildirim2024multimodal}.
However, despite strong performance on benchmarks, our understanding of 
\emph{which} concepts these models have learned---and \emph{how} they use them in the report generation process---remains 
limited. 
This lack of interpretability poses challenges in high-stakes domains like healthcare, where trust, transparency, and safety are critical \cite{quinn2021trust}.

Mechanistic interpretability aims to address some of these challenges by reverse-engineering the internal computations of neural networks \citep{elhage2021mathematical, wang2022interpretability}.
In this context, \acp{SAE} have emerged as a promising tool for inspecting model representations, particularly for \acp{LLM} \citep{bricken2023monosemanticity, cunningham2023sparse}.
By mapping dense model activation vectors to a larger, sparse latent space, it is argued that \acp{SAE} can disentangle human-interpretable `monosemantic' features, which may then be labelled at scale using \acp{LLM}---a process known as automated interpretability.
This approach has enabled insights into model behaviour across various domains, including language \citep{templeton2024scaling, gao_scaling_2024, cunningham2023sparse, lieberum2024gemmascope, he2024llamascope}, vision \citep{stevens2025sparse}, and proteins \citep{adams_mechanistic_2025}.

Given features of interest, we can intervene on the model's internal representations to attempt to control its generations according to them \citep{templeton2024scaling}. 
This provides both a way to validate the discovered concepts and opens up possibilities for fine-grained control \citep{stevens2025sparse, adams_mechanistic_2025, zhang_large_2024}.
In the context of radiology report generation, such steering -- if effective -- could be particularly beneficial, for example: guiding the model to omit pathologies for which its performance is unreliable; encouraging more detailed descriptions of medical devices; or preventing the generation of operational text likely to be fabricated (such as notes about physician communication).

In this work, we explore the feasibility of applying \acp{SAE} to radiology-focused \acp{MLLM}. Specifically, we apply \matsae \citep{bussmann_learning_2025} to \mairatwo~\citep{bannur_maira-2_2024}---one of the most capable publicly available models for grounded and non-grounded \ac{CXR} reporting to date. 
Leveraging large scale, \ac{LLM}-based automated interpretability and scoring methods \citep{paulo_automatically_2024}, we identify a subset of the internal concepts learned by \mairatwo, amongst many features that remain uninterpretable.
These concepts include classical \cxr{}s abnormalities (e.g., atelectasis, cardiomegaly, pleural effusion..), as well as more diverse findings like scoliosis, aortic tortuosity, medical devices, and temporal changes.
We use these discovered concepts in steering experiments, observing mixed success across different scenarios. 
Overall, our findings offer early insights into the internal representations of \mairatwo, marking a step toward greater interpretability. At the same time, this study highlights several practical and methodological challenges across the pipeline---from \ac{SAE} training and automated concept discovery to steering---underscoring the complexity of applying mechanistic interpretability methods to models in specialized domains.
To foster further research in this direction, we publicly release the trained SAE checkpoints, along with the automated feature interpretations\footnote{\url{https://huggingface.co/microsoft/maira-2-sae}}.

\section{Related work}
\label{sec:related_work}

\paragraph{SAE-based interpretability on language models}
\citet{templeton2024scaling} and \citet{gao_scaling_2024} performed automated interpretability studies on large proprietary models, which popularised the use of \acp{SAE} for this task.
\citet{cunningham2023sparse} quantitatively benchmarked the automatic interpretability success of \acp{SAE} against classic unsupervised methods.
Other large-scale interpretability efforts \citep{lieberum2024gemmascope,he2024llamascope} have focused on large public models, training suites of \acp{SAE} across all layers with a range of hyperparameters and publishing their weights and corresponding auto-interpretations, enabling researchers to build on these often-expensive analyses.

\paragraph{SAEs on multimodal language models}
\citet{zhang_large_2024} used \acp{SAE} to analyse visual features in a general-domain multimodal model with architecture similar to \mairatwo, employing a larger multimodal model to automatically explain the learned features.
The \ac{SAE} from \citet{templeton2024scaling} was trained with language data only, then certain features were shown to also respond to related visual inputs.
\citet{lou_sae-v_2025} investigated the potential of \ac{SAE} features to guide data selection for improving modality alignment.
\Acp{SAE} were used by \citet{pach_sparse_2025} to study representations from a vision encoder then steer a downstream language model.
Other classic sparse dictionary learning techniques have also been employed to interpret multimodal language models \citep{parekh2024concept}.

\paragraph{SAEs in biomedical applications}
\citet{le_learning_2024} fitted an \ac{SAE} to a pathology image foundation model, and qualitatively validated several interpretable histological concepts.
\citet{abdulaal2024x} applied \acp{SAE} to a chest X-ray image encoder, then used automatically generated descriptions of \ac{SAE} features to compose radiology reports.
\citet{simon_interplm_2024} and \citet{adams_mechanistic_2025} have also applied \acp{SAE} to interpretation of protein language models, with the former additionally demonstrating steering of protein generation based on interpretable \ac{SAE} features.

\section{Materials and methods}
\begin{figure*}
\centering
    \includegraphics[width=1\textwidth]{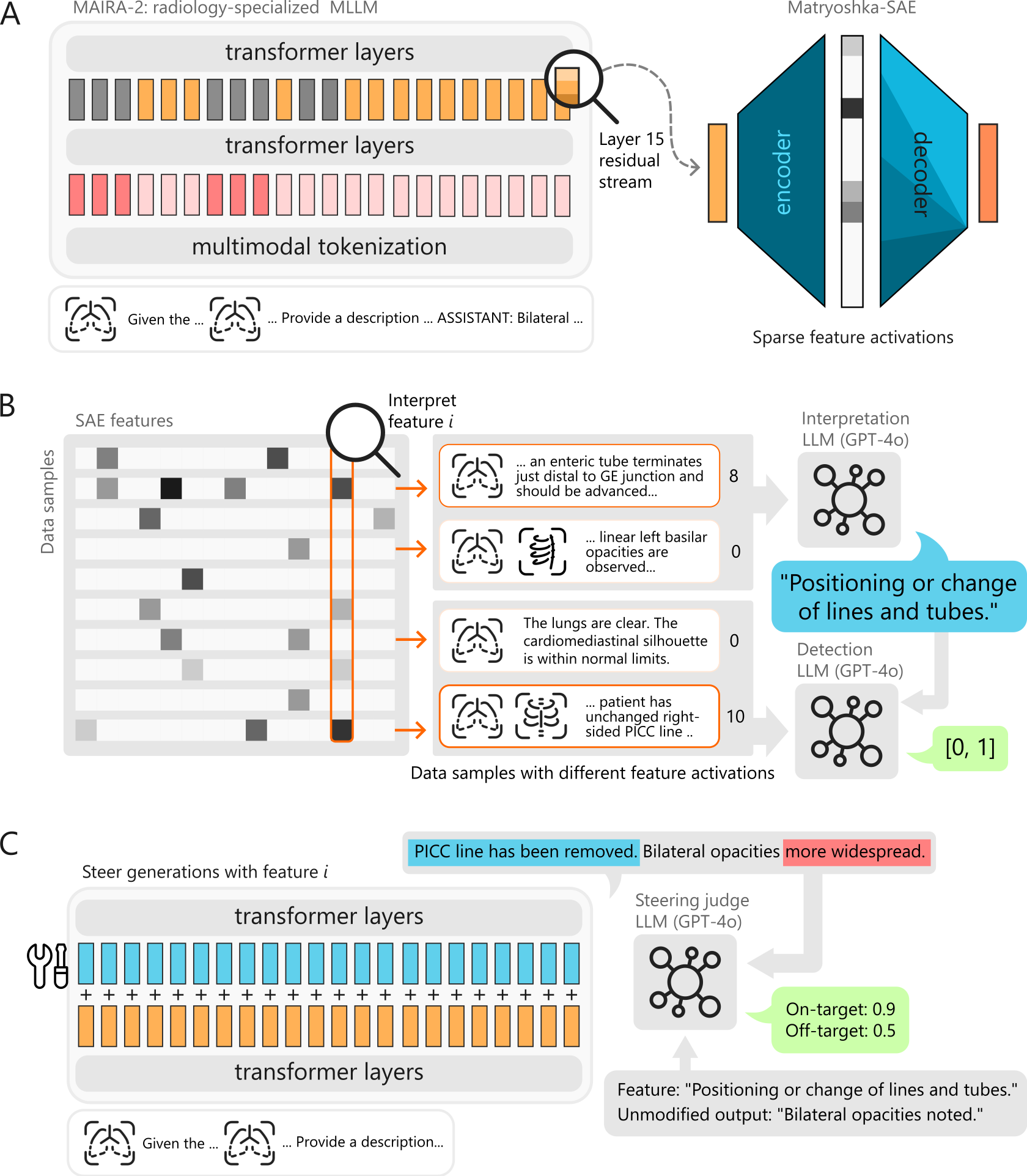}
    \caption{\textbf{Illustration of the study}. \textbf{A:} We train a \matsae using the output of the residual stream of \mairatwo at layer 15. Multimodal tokenization includes obtaining image tokens from an image encoder. We filter out (grey) token indices corresponding to intermediate image tokens and boilerplate parts of the prompt. \textbf{B:} We conduct automated interpretability with \acp{LLM} for both deriving interpretations and scoring interpretations, using detection scoring. \textbf{C:} We try to steer the generations of \mairatwo according to selected features, by adding the corresponding SAE decoder vector to the residual stream at all token positions during each decoding step. Tokenization step omitted for brevity. We evaluate the effectiveness of steering using an \ac{LLM} to judge on-target and off-target effects.}
    \label{fig:illustration}
\end{figure*}

In this section, we describe each element of the experimental pipeline, including the base model and dataset under study, the extraction of internal representations, the SAE used for feature discovery, the procedure for automatically labelling features, and how the feature steering was performed and evaluated. An overview is illustrated in \cref{fig:illustration}.

\subsection{\mairatwo model}
\mairatwo \citep{bannur_maira-2_2024} is a multimodal language model trained for generating the findings section of radiology reports with or without spatial grounding via bounding boxes.
In addition to a frontal \ac{CXR} image, \mairatwo can also optionally ingest: a lateral-view image; indication, technique, and comparison text sections of the current study; and image and report from a prior study of the same patient. This makes it particularly interesting to study since it aligns more closely with real-world clinical workflows than simpler models.

Its architecture is inspired by LLaVA \citep{liu2023llava}.
Each input image is fed through a radiology-specific encoder \citep{perezgarcia2025raddino} and a multilayer perceptron adapter to produce 1369 visual token embeddings.
These are interleaved with text tokens following a custom prompt template (see example in \cref{tab:act-fil}) and given to a 32-layer, 7B-parameter language decoder initialised from Vicuna v1.5 \citep{vicuna2023}.
The adapter and language model were trained on a multi-task mix of CXR datasets totalling 511k multimodal samples from 226k unique patients.

\subsection{Source dataset}
We used \mimiccxr \cite{johnson2019mimic-cxr-dataset}, a subset of \mairatwo's training data, to train \acp{SAE}. This dataset consists of paired de-identified chest radiographs and free-text reports. Focusing on the \emph{findings generation} task, we used 158,555 and 7,906 training and validation samples following the same splits and preprocessing in \citet{bannur_maira-2_2024}. We leveraged \mairatwo's extended input context capabilities to incorporate all available inputs for a given sample.

\subsection{Extraction and filtering of token representations}
We used NNsight \citep{fiotto2024nnsight} for extracting internal representations from the open-source \mairatwo model checkpoint\footnote{ \url{https://huggingface.co/microsoft/maira-2}}.  
We took the hidden states of the residual stream at the output of the middle layer (\#15).
This choice was motivated by prior works that suggest that abstract and semantically rich features are most prominent in the middle layers \citep{templeton2024scaling, lad2024remarkable}.
For each sample, we obtained a sequence of 4096-dimensional token representations, with an average of $3,358 \pm 869$ tokens per sample, reaching a maximum of 5,099 tokens.
This included up to three images---each image being represented by 1,369 tokens---along with additional textual context and the ground-truth findings section.

We refined the extracted representations by removing boilerplate and templated subsequences present in the original prompt from \citet{bannur_maira-2_2024}. This type of filtering is commonly applied when analysing internal representations of instruction-tuned language models \citep{lieberum2024gemmascope}. We extended this approach to the multimodal setting by retaining only the final image token of each \ac{CXR}, to avoid the need to interpret an incomplete image. This filtering procedure resulted in $34.7$M token representations for training, and $1.7$M for validation, substantially reducing the number of tokens to consider. See \cref{tab:act-fil} for an example of token-based representation filtering. In what follows, we refer to these token representations as `samples' as we use them to train and interpret the \ac{SAE}.

\subsection{Sparse autoencoder architecture}
We apply techniques from \emph{sparse dictionary learning} \citep{lee2006efficient, cunningham2023sparse} to disentangle the dense representations of \mairatwo into potentially-interpretable sparse features.
Specifically, we employ \matsae~\citep{bussmann_learning_2025}.
By simultaneously training multiple nested autoencoders of increasing size, \matsae is argued to reduce feature splitting and produce more interpretable features.

Formally, given a total dictionary size $m$ and an input $\mathbf{x} \in \mathbb{R}^n$, the \matsae training objective encourages good reconstruction at multiple levels at the same time:
\begin{gather}
\mathcal{L}(\mathbf{x}) = \sum_{m_j \in \mathcal{M}} \lVert \mathbf{x} - 
\underbrace{( \mathbf{W}^{\mathrm{dec}}_{1:m_j} \mathbf{f}(\mathbf{x})_{1:m_j} + \mathbf{b}^{\mathrm{dec}} )}_{\mathclap{\text{reconstruction using first $m_j$ features}}} 
\rVert_2^2 + \alpha \mathcal{L}_{\mathrm{aux}} \\
\mathbf{f}(\mathbf{x}) = \sigma(\mathbf{W}^{\mathrm{enc}}\mathbf{x} + \mathbf{b}^{\mathrm{enc}}) \,,
\end{gather}
where ${1:m_j}$ denotes taking the first $m_j$ elements, $\mathcal{M}$ is the set of nested dictionaries of increasing sizes $m_1< m_2<...<{m_{|\mathcal{M}|}=m}$. $\mathbf{W}^{\mathrm{enc}}\in \mathbb{R}^{m \times n}, \mathbf{b}^{\mathrm{enc}} \in \mathbb{R}^m$ are the encoder matrix and bias, $\mathbf{W}^{\mathrm{dec}} \in \mathbb{R}^{n \times m}, \mathbf{b}^{\mathrm{dec}} \in \mathbb{R}^n$ are the decoder matrix and bias. Similar to 
\citet{bussmann_learning_2025}, we use BatchTopK \citep{bussmann2024batchtopk} as activation function $\sigma$ (with slight abuse of notation), which builds on TopK \citep{gao_scaling_2024} to enforce an average sparsity of $k$ active features across a batch of size $b$.
Finally, $\mathcal{L}_{\mathrm{aux}}$ is the auxiliary loss suggested in \citet{gao_scaling_2024} to reduce dead neurons, further described in \cref{sec:app:aux-loss}. In what follows, \feature{i} refers to the $i$'th feature in the sparse latent space of the SAE, with $i \in [0, m-1]$.

\subsection{SAE training}
We relied on the open-source \texttt{dictionary\_learning} repository\footnote{\url{https://github.com/saprmarks/dictionary_learning}} \citep{marks2024dictionary_learning} to train the \acp{SAE} used in this study. We rescaled all token representations similarly to \citet{gao_scaling_2024}, with a normalization factor of $22.34$, computed from the full training set as the mean $\ell_2$ norm of the representations. We benchmarked several SAE architectures, including \topk~\citep{gao_scaling_2024}, \batchtopk~\citep{bussmann2024batchtopk}, and \matsae~\citep{bussmann_learning_2025}, and found that the latter achieved the best reconstruction performance in this context.
The width of \acp{SAE} is commonly specified as a factor of the hidden dimensionality of the base model (4096 for \mairatwo).
We experimented with various expansion factors $\mathrm{ef} \in \{2, 4, 8, 16\}$. Although ${\mathrm{ef}=16}$ achieved the best reconstruction performance, we selected ${\mathrm{ef}=4}$ as a trade-off between reconstruction quality, sparsity, and the computational cost of the downstream auto-interpretation pipeline that requires extensive LLM API calls.  We also tested multiple values of the sparsity parameter $k \in \{32, 64, 128, 256\}$, with ${k=256}$ producing the most favourable results in terms of reconstruction fidelity and lowest number of dead features, consistent with the settings in \citet{zhang_large_2024}. Unless stated otherwise, all downstream results are reported using \matsae with ${\mathrm{ef}=4}$ and  ${k=256}$, resulting in \nlatents features. A full list of hyperparameters is provided in \cref{tab:sae-hparams}.

\subsection{Automated interpretation of SAE features}
To identify which \ac{SAE} features correspond to human-interpretable concepts, we follow prior work in interpreting features at scale with an automated, \ac{LLM}-based pipeline \citep{paulo_automatically_2024,bricken2023monosemanticity,bills2023language}. 

For each feature, we randomly collect samples that do not activate it ($f_i(\mathbf{x})=0$) and an equal number with activation strength in the top observed decile. We found this sampling strategy superior to the fully stratified sampling done in earlier work~\citep{paulo_automatically_2024}. We then query an \ac{LLM} (GPT-4o version 
2024-11-20) to produce an interpretation of the observed activation patterns.
Although \mairatwo consumes interleaved images and text, we provide our interpretation \ac{LLM} with text inputs only, because the reliability of the generalist GPT-4* series on \ac{CXR} image understanding has been questioned~\citep{Yan2024WorseTR,Jiang2024GPT4VCG}.

We score the quality of these interpretations following the detection approach from \citet{paulo_automatically_2024}, which scores a (textual) feature interpretation by how well it can be used to predict whether that feature will be active on a new sample, again using an \ac{LLM}. Where possible, we balance the number of positive and negative samples\footnote{It was not always possible to find 100 positive examples for a given feature, resulting in an imbalanced evaluation set. However, we did not observe a correlation between imbalance and \fone.} and report detection \fone on the evaluation set as a measure of feature interpretability. For scoring we sample from the top quintile and again hold non-activating samples as the balanced `negatives'.

We use 50 data examples per feature to derive an interpretation, and 200 (non-overlapping) examples to score interpretation quality. We source these examples from a set of 500,000 randomly selected samples from the dataset used to train the SAE. We generate and score interpretations for every feature in the \ac{SAE} for which it is possible to collect samples as prescribed above, resulting in interpretations for 99.5\% (16,299) of the \nlatents features.
Further details on automated interpretation, including \ac{LLM} prompts, are in \Cref{sec:app:auto_interp,sec:app:interp_scoring}.

\subsection{Steering}
\ac{SAE} features can be used to manipulate a model's output towards or away from desired concepts in a process called steering~\citep{templeton2024scaling, adams_mechanistic_2025, zhang_large_2024}.
To steer model generations using a single SAE feature $f_i$, we extract the corresponding column $\mathbf{W}_i^\mathrm{dec}$ from the SAE decoder matrix as a steering vector. 
This vector is multiplied by a coefficient $\alpha$ and added at each decoding step to the hidden states for every token in the sequence---whether in the prompt or generated tokens \citep{turner2023steering, li2023inferencetime, durmus2024steering}.
By intervening on the hidden states in this manner, we expect the model generations to introduce or emphasise content associated with the targeted concept, reflecting an induced high activation of that feature.
In contrast, applying negative coefficients is expected to de-emphasise, remove, or even invert the corresponding concept in the model generation.

Finding the right steering coefficient $\alpha$ is non-trivial. After initial experiments with a wide range of steering coefficients ($\alpha \in \{1, 2.5, 5, 10, 15, 25, 50, 100\}$), we chose $\alpha=10$ for positive and $\alpha=-10$ for negative steering as they strike a good balance between the appearance of noticeable targeted changes while remaining in-distribution.

\subsection{Evaluating steering success}\label{sec:eval_steering_success}
Successful steering should modify only the concept of interest while keeping the rest unchanged. We therefore evaluate the effectiveness of steering using two dimensions: \textit{on-target} changes refer to modifications directly related to the concept of interest, while \textit{off-target} changes are all other significant alterations of the content of the generated report.
Both dimensions of change are measured in comparison between the reports generated with and without the steering intervention.
To enable large-scale evaluation of steering success of multiple concepts, we use an \ac{LLM}-as-a-judge approach similar to \citet{wu_axbench_2025}. The \ac{LLM} is prompted to output on-target and off-target scores from 0 to 1, %
where higher scores correspond to stronger changes. 
More details about the LLM judge are in \cref{sec:app:steering_evaluation}.

\section{Findings}

\begin{figure}
    \centering
    \includegraphics[width=0.5\textwidth]{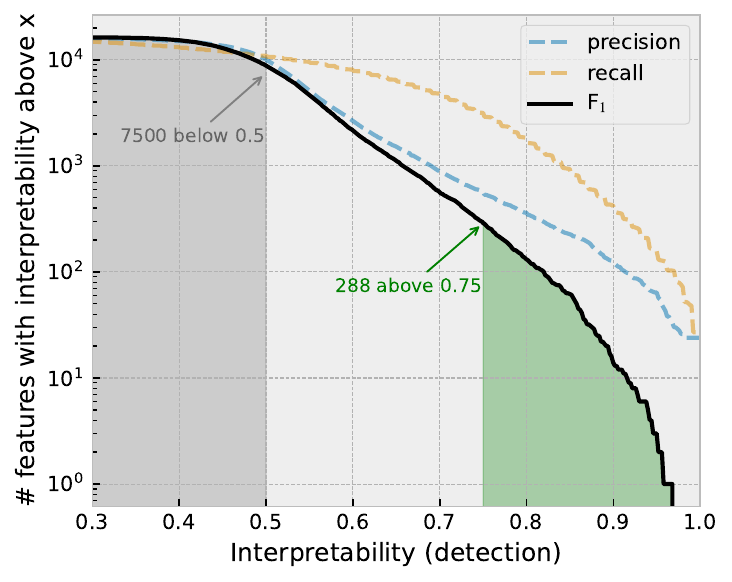}
    \caption{\textbf{Interpretable features exist, but are rare.} Using detection \fone on the evaluation set as a measure of interpretability, we show that from \nlatents features, 288 (1.8\%) score above 0.75, whereas 7,500 (46\%) score below 0.5 (random performance for a balanced evaluation set).}
    \label{fig:latents_interpretability}
\end{figure}

\subsection{\mairatwo contains interpretable features}
\Cref{fig:latents_interpretability} shows the distribution of detection \fone across the \nlatents features in our \ac{SAE}. With 7,500 (46\%) features below 0.5, this illustrates that many features are not interpretable by our automated pipeline. There do however exist consistently interpretable features, with 288 (1.8\%) exhibiting \fone above 0.75. We observe consistently higher recall than precision, indicating that feature descriptions tend towards being non-specific.

We observe that highly interpretable features tend to pertain to the presence of abnormal findings (\feature{1336}, $\fone=0.89$: \featexpl{Aortic tortuosity or calcification identified in chest imaging.}), changes especially of medical lines and tubes (\feature{11240}, $\fone=0.96$: \featexpl{Descriptions of findings related to chest tube placement or removal.}), and reporting style (\feature{12106}, $\fone=0.93$: \featexpl{Use of `however' in clinical findings indicating possible issues needing further investigation.}). 

Features with low detection \fone are most commonly ascribed generic descriptions of the task of radiology reporting emphasising comparison with prior imaging (\feature{730} and 16 others, $\fone=0.50$: \featexpl{Comparative analysis of current and prior imaging findings.}, \feature{12117} and 5 others, $\fone=0.40$, \featexpl{Comparison of current imaging findings to prior studies.}), and less frequently references to the instruction given to \mairatwo (\feature{2653}, $\fone=0.42$, \featexpl{Provide findings description in comparison with prior images.}), or pathological changes (\feature{10005}, $\fone=0.43$, \featexpl{Enlargement of cardiac silhouette often with pulmonary vascular changes.}).

Even among highly interpretable features, we observe a degree of repetition in feature descriptions: of the 61 features with \fone above 0.85, we observe seven instances of apparent repetition. For example, \feature{1374} and \feature{6105} are \featexpl{Presence of atelectasis in imaging findings.} and \featexpl{Mentions and descriptions of atelectasis} respectively, and \feature{13515} and \feature{13911} are both described as \featexpl{Elevation of the hemidiaphragm}.

\begin{figure*}[t]
    \centering
    \includegraphics[width=\textwidth]{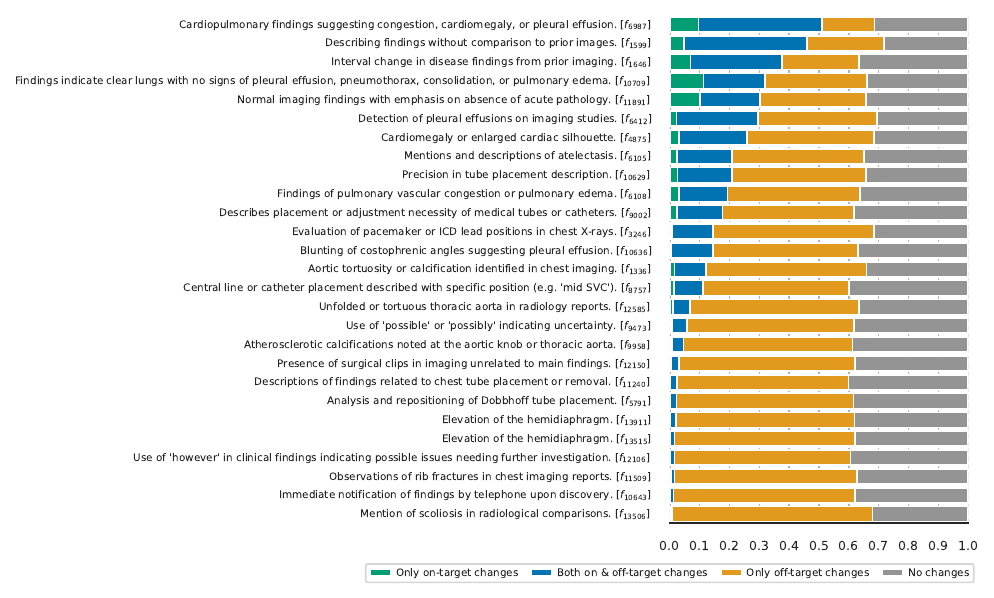}
    \caption{On-target and off-target effects of steering features \feature{i} with $\alpha=10$. The scores of the LLM judge are binarized at $0.1$ and the bars show the proportion of samples where steering led only to on-target changes, only to off-target changes, both, or none at all. Results are shown for the full validation set including 7,906 studies.}
    \label{fig:steering}
\end{figure*}

\subsection{Steering success depends on the feature}\label{sec:findings_steering}
\begin{table*}[t]
\centering
\caption{Examples of steered reports including LLM-based evaluation. Following the colouring from \Cref{fig:steering}, on-target changes between the original and the steered report are highlighted in \textcolor{ForestGreen}{green}, and off-target effects in \textcolor{Orange}{orange}.}
\label{tab:steering-examples}
\footnotesize
\begin{tabular}{@{}p{0.06\linewidth} p{0.91\linewidth}@{}}
\toprule
\multicolumn{2}{c}{\textcolor{ForestGreen}{\bf Only on-target changes}}  \\
\midrule
\textbf{Feature} & \feature{1599}, $\fone=0.79$: \featexpl{Describing findings without comparison to prior images.} \\
\cmidrule{2-2}
\textbf{Original} & \textcolor{ForestGreen}{Compared with the prior study, lung volumes are lower}, causing bronchovascular crowding. However, no focal consolidation, pleural effusion, or pneumothorax. \textcolor{ForestGreen}{Cardiomediastinal and hilar silhouettes are unchanged}. \\
\cmidrule{2-2}
\textbf{Steered} $\balpha~\mathbf{=10}$ & \textcolor{ForestGreen}{Lung volumes are low}, causing bronchovascular crowding. No focal consolidation, pleural effusion, or pneumothorax. \textcolor{ForestGreen}{Cardiomediastinal and hilar silhouettes are unremarkable}. \\
\cmidrule{2-2}
\textbf{Scores} & on-target score=1.0 off-target score=0.0 \\
\midrule[\heavyrulewidth]
\multicolumn{2}{c}{\textcolor{RoyalBlue}{\bf Both on- and off-target changes}}  \\
\midrule
\textbf{Feature} & \feature{6412}, $\fone=0.84$: \featexpl{Detection of pleural effusions on imaging studies.} \\
\cmidrule{2-2}
\textbf{Original} & The heart size is normal. The mediastinal and hilar contours are unremarkable. The pulmonary vascularity is normal. Patchy ill-defined opacity is noted within the left lung base, which may represent an area of developing infection. \textcolor{ForestGreen}{No pleural effusion} or pneumothorax is present. There are no acute osseous abnormalities. \\
\cmidrule{2-2}
\textbf{Steered} $\balpha~\mathbf{=10}$ & The heart size is normal. The mediastinal and hilar contours are unremarkable. The pulmonary vascularity is normal. \textcolor{ForestGreen}{There is a small left pleural effusion}. \textcolor{Orange}{Minimal streaky opacity in the left lung base may reflect atelectasis, though infection cannot be completely excluded.} There is no pneumothorax. No acute osseous abnormality is seen. \\
\cmidrule{2-2}
\textbf{Scores} & on-target score=1.0 off-target score=0.2 \\
\midrule[\heavyrulewidth]
\multicolumn{2}{c}{\textcolor{Orange}{\bf Only off-target changes}}  \\
\midrule
\textbf{Feature} & \feature{10643}, $\fone=0.86$: \featexpl{Immediate notification of findings by telephone upon discovery.} \\
\cmidrule{2-2}
\textbf{Original} & Since \_, small right pleural effusion is unchanged, right pleural catheter is in unchanged position. The lungs are clear. The cardiomediastinal silhouette, hilar contours, and pleural surfaces are normal. No pneumothorax. \\
\cmidrule{2-2}
\textbf{Steered} $\balpha~\mathbf{=10}$ & Since \_, a right pleural drainage catheter is in place. \textcolor{Orange}{A small right apical pneumothorax is seen.} A small right pleural effusion is seen. \textcolor{Orange}{A left pleural effusion is small.} The lungs are clear. The heart size is normal. \textcolor{Orange}{Tips in the esophagus are noted.} \\
\cmidrule{2-2}
\textbf{Scores} & on-target score=0.1 off-target score=0.7 \\
\bottomrule

\end{tabular}
\end{table*}

In this section, we evaluate to what extent the \ac{SAE} features can be used to effectively steer the model generation into desired directions.
\Cref{fig:steering} shows the steering performance across a set of highly interpretable features on the full validation set of 7,906 \ac{CXR} studies. The LLM-based steering evaluation produces two metrics: an \textit{on-target score} and an \textit{off-target score}.
Based on these metrics, we stratify the results into four disjoint cases: (1)~only on-target changes were observed; (2)~both on-target and off-target changes; (3)~only off-target changes; and (4)~no observable changes occurred following steering.
We show examples from each category in \cref{tab:steering-examples}.
We describe the process of selecting features for steering and additional results and examples for both positive and negative steering in \Cref{app:feature_steering_details}.

Overall, feature steering produces more off-target changes than on-target changes.
Purely on-target changes are relatively rare, with the highest observed proportion being $11.3\%$, seen in \feature{10709}.
An example of such a case is shown for \feature{1599} in the top row of \cref{tab:steering-examples}, where the steered generation effectively removed all comparisons to prior studies.

More commonly, on-target effects are accompanied by varying degrees of off-target changes.
In \feature{6412}, for example, $27.4\%$ of cases exhibit both on- and off-target changes.
While some of these off-target changes reflect underlying correlations in the dataset---for example in \feature{6412} in \cref{tab:steering-examples}, where the output includes \textit{atelectasis} alongside \textit{pleural effusion}, consistent with known clinical correlations---most off-target changes are not due to such causal relationships. In many cases, off-target changes are seen when the original report is non-specific (For example, ``No significant interval change'') while the steered generation is more descriptive, leading to large off-target changes due to the discrepancy. 

In most cases, steering leads predominantly to off-target effects.
Extreme cases include, but are not limited to, \feature{11509}, \feature{10643}, and \feature{13506} where more than 50\% of the steered samples contain off-target changes despite showing almost no on-target changes.
The corresponding example for \feature{10643}  in \cref{tab:steering-examples} adds a right apical pneumothorax, a left pleural effusion, and tips descriptions, none of which are directly related to the feature's description of \featexpl{Immediate notifications of findings by telephone upon discovery.}

In approximately $35\%$ of cases across all features, steering leads to no observable changes. 

Interestingly, we observe a strong correlation in the on-target effects between steering features with a positive and with a negative coefficient $\alpha$ (Spearman's $\rho = 0.90$, $p \leq 0.05$, $n = 67$). The correlation is also present for off-target effects, although less pronounced (Spearman's $\rho = 0.71$, $p \leq 0.05$, $n = 67$).
We also observe a correlation between the activation frequency of a feature and its on-target score (Spearman's $\rho = 0.40$, $p \leq 0.05$, $n = 50$) for features with $\fone > 0.85$. We do not observe a significant correlation of the activation frequency with off-target scores. 

This section leaves us with two main observations: (1) the success of feature steering highly depends on the selected feature and the steered report, and (2) even when successful, feature steering regularly produces more off-target effects than on-target effects.

\section{Discussion and conclusion}
Motivated by the success of \ac{SAE}-based approaches for the interrogation of \acp{LLM}, in this work we investigated the radiology-specialized \ac{MLLM} \citet[\mairatwo;][]{bannur_maira-2_2024} using \matsae \citep{bussmann_learning_2025}. We based our automated interpretability and scoring pipeline on an established approach \citep{paulo_automatically_2024}, with some modifications specific to the radiology domain. We then explored the extent to which \mairatwo can be steered using the discovered interpretable features.

We found a small set of highly interpretable \ac{SAE} features, representing a variety of concepts pertinent to radiology reporting, and often with more granularity than existing concept categories such as the CheXpert classes~\cite{irvin2019chexpert}. Among these highly interpretable features, we observed some repeated or similar feature descriptions, which may be evidence for feature composition~\cite{anders2024sparse, wattenberg2024relationalcompositionneuralnetworks} or splitting~\cite{bricken2023monosemanticity}.
Almost half of the \ac{SAE} features were found not to be interpretable by our pipeline. It is difficult to establish the base rate of `interpretable' features discovered in \ac{SAE} analyses as most prior work reports on subsets of features~\citep{paulo_automatically_2024,bussmann_learning_2025,Minder2025RobustlyIC,templeton2024scaling} or subsets of evaluation samples~\citep{mudide2024efficient}, whereas here we report scores for almost all \ac{SAE} features.

Nonetheless, it may be possible to uncover more interpretable features in \mairatwo with improvements in automated interpretability, for example making use of a radiology-adept \ac{MLLM} to more comprehensively capture imaging content, by further improving the selection of dataset exemplars (e.g.~using `hard' negatives as in \citet{Minder2025RobustlyIC}), or by actively seeking `output' features \citep{paulo_automatically_2024, gur2025enhancing} through an intervention-based interpretation pipeline~\citep{shaham2024multimodal}. 
However, given our specialised domain, we are limited in the set of models available for e.g.~\ac{CXR} visual question-answering, or generation of counterfactual examples.

Our study discovered two recurring patterns related to steering success.
First, we found a significant correlation between the activation frequency of features in the training dataset and steering success for highly interpretable features.
Given that the training data in this study was a subset of \mairatwo's training data, we hypothesise that frequently activating features are well learned and better represented in the language model’s latent space---which in return is the cause for the high steering success of such features.
Second, features that are highly steerable in the positive direction also show good steering performance in the negative direction and vice versa.
This suggests that the learned features retain their approximate linearity far below the point where they usually activate.
Other recent studies on steering \acp{LLM} with hand-picked \ac{SAE} features also found disproportionately many failure cases, but could not provide an empirical explanation \citep{durmus2024steering, obrien2024steering}.
This also relates to the comprehensive steering benchmarking study from \citet{wu_axbench_2025}, wherein almost all steering methods (including \ac{SAE}-based) dramatically underperformed prompting and fine-tuning.

Alongside low steering success, we also observed adverse effects from steering, causing sometimes clinically significant changes unrelated to the concept of interest, such as the confabulation or omission of findings.
Compared to generalist models used by \citet{durmus2024steering} and \citet{obrien2024steering}, the nature of our application both required and allowed us to perform a more principled evaluation of the side-effects of steering than these works, which revealed that these off-target changes often exceed those observed along the concept dimension.

The reasons for the low steering success and the large adverse effects remain speculative but could be rooted in limitations of the \acp{SAE} used for concept discovery, of the steering method, or of the automated feature interpretation and steering evaluation pipelines.
For example, quantitatively measuring on-target effects is challenging when a concept is already present.
This is, for example, observed for \feature{6412}: attempting to add more pleural effusions to a report that already describes bilateral pleural effusions should rightfully result in no change compared to the original output.
This effect is especially relevant in negative steering since most concepts are already absent in the majority of samples.
More fundamentally, it is possible that the \ac{SAE} features themselves are not well disentangled \citep[cf.~feature splitting/absorption/composition;][]{bricken2023monosemanticity, chanin_is_2024, anders2024sparse}, or that representations of concepts of interest are significantly nonlinear in \mairatwo \citep{park2024linear}.
Some of the problems above could be alleviated by more informed steering methods that would especially incorporate knowledge about \emph{which} tokens to modify and \emph{how much}.

In conclusion, this study provides early insights into the inner workings of a radiology-specialised \ac{MLLM}, while also highlighting various practical and technical challenges in applying mechanistic interpretability methods to domain-specific models. Notably:
\begin{itemize}
    \item We identify a subset of the internal concepts in \mairatwo that are often more fine-grained than existing findings labels for \ac{CXR} reports, despite many features remaining uninterpretable.
    \item While concept steering shows potential, its effectiveness varies considerably by feature and case---highlighting the need for further investigation.
    \item These findings lay a foundation for more transparent, interpretable, and controllable radiology \acp{MLLM}, and support continued research through the planned release of the SAE checkpoint and automated interpretations.
\end{itemize}
By releasing\footnote{ \url{https://huggingface.co/microsoft/maira-2-sae}} the trained \acp{SAE} and LLM-generated interpretations for all features, we hope to facilitate further research into the application of mechanistic interpretability to specialised multimodal models such as \mairatwo.

\section*{Impact Statement}
This paper focuses on approaches to understand the internal representations of a radiology-adapted \ac{MLLM} and potentially control its behaviour through steering. The use of \ac{MLLM} on medical data can have positive or negative impacts depending on the context of use since such models can generate incorrect or misleading outputs. Extracting interpretable features from model internals may provide model builders a route towards designing safer or more effective models. Steering could provide an additional mechanism by which models can be controlled, either to enhance or suppress unsafe or other undesirable behaviours.

\bibliography{bibliography}

\begin{thebibliography}{55}
\providecommand{\natexlab}[1]{#1}
\providecommand{\url}[1]{\texttt{#1}}
\expandafter\ifx\csname urlstyle\endcsname\relax
  \providecommand{\doi}[1]{doi: #1}\else
  \providecommand{\doi}{doi: \begingroup \urlstyle{rm}\Url}\fi

\bibitem[Abdulaal et~al.(2024)Abdulaal, Fry, Monta{\~n}a-Brown, Ijishakin, Gao, Hyland, Alexander, and Castro]{abdulaal2024x}
Abdulaal, A., Fry, H., Monta{\~n}a-Brown, N., Ijishakin, A., Gao, J., Hyland, S., Alexander, D.~C., and Castro, D.~C.
\newblock An {X}-ray is worth 15 features: Sparse autoencoders for interpretable radiology report generation.
\newblock \emph{arXiv preprint arXiv:2410.03334}, 2024.

\bibitem[Adams et~al.(2025)Adams, Bai, Lee, Yu, and AlQuraishi]{adams_mechanistic_2025}
Adams, E., Bai, L., Lee, M., Yu, Y., and AlQuraishi, M.
\newblock From {Mechanistic} {Interpretability} to {Mechanistic} {Biology}: {Training}, {Evaluating}, and {Interpreting} {Sparse} {Autoencoders} on {Protein} {Language} {Models}, February 2025.
\newblock URL \url{https://www.biorxiv.org/content/10.1101/2025.02.06.636901v1}.
\newblock Pages: 2025.02.06.636901 Section: New Results.

\bibitem[Anders et~al.(2024)Anders, Neo, Hoelscher-Obermaier, and Howard]{anders2024sparse}
Anders, E., Neo, C., Hoelscher-Obermaier, J., and Howard, J.~N.
\newblock Sparse autoencoders find composed features in small toy models, 2024.
\newblock URL \url{https://www.lesswrong.com/posts/a5wwqza2cY3W7L9cj}.

\bibitem[Bannur et~al.(2024)Bannur, Bouzid, Castro, Schwaighofer, Bond-Taylor, Ilse, Pérez-García, Salvatelli, Sharma, Meissen, Ranjit, Srivastav, Gong, Falck, Oktay, Thieme, Lungren, Wetscherek, Alvarez-Valle, and Hyland]{bannur_maira-2_2024}
Bannur, S., Bouzid, K., Castro, D.~C., Schwaighofer, A., Bond-Taylor, S., Ilse, M., Pérez-García, F., Salvatelli, V., Sharma, H., Meissen, F., Ranjit, M., Srivastav, S., Gong, J., Falck, F., Oktay, O., Thieme, A., Lungren, M.~P., Wetscherek, M.~T., Alvarez-Valle, J., and Hyland, S.~L.
\newblock {MAIRA}-2: {Grounded} {Radiology} {Report} {Generation}, June 2024.
\newblock URL \url{http://arxiv.org/abs/2406.04449}.

\bibitem[Bills et~al.(2023)Bills, Cammarata, Mossing, Tillman, Gao, Goh, Sutskever, Leike, Wu, and Saunders]{bills2023language}
Bills, S., Cammarata, N., Mossing, D., Tillman, H., Gao, L., Goh, G., Sutskever, I., Leike, J., Wu, J., and Saunders, W.
\newblock Language models can explain neurons in language models.
\newblock \url{https://openaipublic.blob.core.windows.net/neuron-explainer/paper/index.html}, 2023.

\bibitem[Bricken et~al.(2023)Bricken, Templeton, Batson, Chen, Jermyn, Conerly, Turner, Anil, Denison, Askell, Lasenby, Wu, Kravec, Schiefer, Maxwell, Joseph, Hatfield-Dodds, Tamkin, Nguyen, McLean, Burke, Hume, Carter, Henighan, and Olah]{bricken2023monosemanticity}
Bricken, T., Templeton, A., Batson, J., Chen, B., Jermyn, A., Conerly, T., Turner, N., Anil, C., Denison, C., Askell, A., Lasenby, R., Wu, Y., Kravec, S., Schiefer, N., Maxwell, T., Joseph, N., Hatfield-Dodds, Z., Tamkin, A., Nguyen, K., McLean, B., Burke, J.~E., Hume, T., Carter, S., Henighan, T., and Olah, C.
\newblock Towards monosemanticity: Decomposing language models with dictionary learning.
\newblock \emph{Transformer Circuits Thread}, 2023.
\newblock https://transformer-circuits.pub/2023/monosemantic-features/index.html.

\bibitem[Bussmann et~al.(2024)Bussmann, Leask, and Nanda]{bussmann2024batchtopk}
Bussmann, B., Leask, P., and Nanda, N.
\newblock {BatchTopK} sparse autoencoders.
\newblock \emph{arXiv preprint arXiv:2412.06410}, 2024.

\bibitem[Bussmann et~al.(2025)Bussmann, Nabeshima, Karvonen, and Nanda]{bussmann_learning_2025}
Bussmann, B., Nabeshima, N., Karvonen, A., and Nanda, N.
\newblock Learning {Multi}-{Level} {Features} with {Matryoshka} {Sparse} {Autoencoders}, March 2025.
\newblock URL \url{http://arxiv.org/abs/2503.17547}.
\newblock arXiv:2503.17547 [cs].

\bibitem[Chanin et~al.(2024)Chanin, {Wilken-Smith}, Dulka, Bhatnagar, and Bloom]{chanin_is_2024}
Chanin, D., {Wilken-Smith}, J., Dulka, T., Bhatnagar, H., and Bloom, J.~I.
\newblock {A} is for absorption: Studying feature splitting and absorption in sparse autoencoders.
\newblock In \emph{NeurIPS 2024 Workshop on Interpretable AI}, December 2024.
\newblock URL \url{https://openreview.net/forum?id=Wzav8fesTL}.

\bibitem[Chen et~al.(2024)Chen, Varma, Delbrouck, Paschali, Blankemeier, Van~Veen, Valanarasu, Youssef, Cohen, Reis, et~al.]{chen2024chexagent}
Chen, Z., Varma, M., Delbrouck, J.-B., Paschali, M., Blankemeier, L., Van~Veen, D., Valanarasu, J. M.~J., Youssef, A., Cohen, J.~P., Reis, E.~P., et~al.
\newblock {CheXagent}: Towards a foundation model for chest x-ray interpretation.
\newblock \emph{arXiv preprint arXiv:2401.12208}, 2024.

\bibitem[Chiang et~al.(2023)Chiang, Li, Lin, Sheng, Wu, Zhang, Zheng, Zhuang, Zhuang, Gonzalez, Stoica, and Xing]{vicuna2023}
Chiang, W.-L., Li, Z., Lin, Z., Sheng, Y., Wu, Z., Zhang, H., Zheng, L., Zhuang, S., Zhuang, Y., Gonzalez, J.~E., Stoica, I., and Xing, E.~P.
\newblock Vicuna: An open-source chatbot impressing {GPT-4} with 90\%* {ChatGPT} quality, March 2023.
\newblock URL \url{https://lmsys.org/blog/2023-03-30-vicuna/}.

\bibitem[Cunningham et~al.(2024)Cunningham, Ewart, Riggs, Huben, and Sharkey]{cunningham2023sparse}
Cunningham, H., Ewart, A., Riggs, L., Huben, R., and Sharkey, L.
\newblock Sparse autoencoders find highly interpretable features in language models.
\newblock In \emph{The Twelfth International Conference on Learning Representations}, 2024.
\newblock URL \url{https://openreview.net/forum?id=F76bwRSLeK}.

\bibitem[Durmus et~al.(2024)Durmus, Tamkin, Clark, Wei, Marcus, Batson, Handa, Lovitt, Tong, McCain, Rausch, Huang, Bowman, Ritchie, Henighan, and Ganguli]{durmus2024steering}
Durmus, E., Tamkin, A., Clark, J., Wei, J., Marcus, J., Batson, J., Handa, K., Lovitt, L., Tong, M., McCain, M., Rausch, O., Huang, S., Bowman, S., Ritchie, S., Henighan, T., and Ganguli, D.
\newblock Evaluating feature steering: A case study in mitigating social biases, 2024.
\newblock URL \url{https://anthropic.com/research/evaluating-feature-steering}.

\bibitem[Elhage et~al.(2021)Elhage, Nanda, Olsson, Henighan, Joseph, Mann, Askell, Bai, Chen, Conerly, et~al.]{elhage2021mathematical}
Elhage, N., Nanda, N., Olsson, C., Henighan, T., Joseph, N., Mann, B., Askell, A., Bai, Y., Chen, A., Conerly, T., et~al.
\newblock A mathematical framework for transformer circuits.
\newblock \emph{Transformer Circuits Thread}, 2021.

\bibitem[Fiotto-Kaufman et~al.(2024)Fiotto-Kaufman, Loftus, Todd, Brinkmann, Juang, Pal, Rager, Mueller, Marks, Sharma, et~al.]{fiotto2024nnsight}
Fiotto-Kaufman, J., Loftus, A.~R., Todd, E., Brinkmann, J., Juang, C., Pal, K., Rager, C., Mueller, A., Marks, S., Sharma, A.~S., et~al.
\newblock {NNsight} and {NDIF}: Democratizing access to foundation model internals.
\newblock \emph{arXiv preprint arXiv:2407.14561}, 2024.

\bibitem[Gao et~al.(2024)Gao, la~Tour, Tillman, Goh, Troll, Radford, Sutskever, Leike, and Wu]{gao_scaling_2024}
Gao, L., la~Tour, T.~D., Tillman, H., Goh, G., Troll, R., Radford, A., Sutskever, I., Leike, J., and Wu, J.
\newblock Scaling and evaluating sparse autoencoders, June 2024.
\newblock URL \url{http://arxiv.org/abs/2406.04093}.
\newblock arXiv:2406.04093 [cs].

\bibitem[Gur-Arieh et~al.(2025)Gur-Arieh, Mayan, Agassy, Geiger, and Geva]{gur2025enhancing}
Gur-Arieh, Y., Mayan, R., Agassy, C., Geiger, A., and Geva, M.
\newblock Enhancing automated interpretability with output-centric feature descriptions.
\newblock \emph{arXiv preprint arXiv:2501.08319}, 2025.

\bibitem[He et~al.(2024)He, Shu, Ge, Chen, Wang, Zhou, Liu, Guo, Huang, Wu, Jiang, and Qiu]{he2024llamascope}
He, Z., Shu, W., Ge, X., Chen, L., Wang, J., Zhou, Y., Liu, F., Guo, Q., Huang, X., Wu, Z., Jiang, Y.-G., and Qiu, X.
\newblock {L}lama {S}cope: Extracting millions of features from {Llama-3.1-8B} with sparse autoencoders, 2024.
\newblock URL \url{https://arxiv.org/abs/2410.20526}.

\bibitem[Hyland et~al.(2023)Hyland, Bannur, Bouzid, Castro, Ranjit, Schwaighofer, P{\'e}rez-Garc{\'\i}a, Salvatelli, Srivastav, Thieme, et~al.]{hyland2023maira}
Hyland, S.~L., Bannur, S., Bouzid, K., Castro, D.~C., Ranjit, M., Schwaighofer, A., P{\'e}rez-Garc{\'\i}a, F., Salvatelli, V., Srivastav, S., Thieme, A., et~al.
\newblock {MAIRA}-1: A specialised large multimodal model for radiology report generation.
\newblock \emph{arXiv:2311.13668}, 2023.
\newblock URL \url{https://arxiv.org/abs/2311.13668}.

\bibitem[Irvin et~al.(2019)Irvin, Rajpurkar, Ko, Yu, Ciurea-Ilcus, Chute, Marklund, Haghgoo, Ball, Shpanskaya, et~al.]{irvin2019chexpert}
Irvin, J., Rajpurkar, P., Ko, M., Yu, Y., Ciurea-Ilcus, S., Chute, C., Marklund, H., Haghgoo, B., Ball, R., Shpanskaya, K., et~al.
\newblock {CheXpert}: A large chest radiograph dataset with uncertainty labels and expert comparison.
\newblock In \emph{Proceedings of the AAAI conference on artificial intelligence}, volume~33, pp.\  590--597, 2019.

\bibitem[Jiang et~al.(2024)Jiang, Chen, Nguyen, Mervak, and Tan]{Jiang2024GPT4VCG}
Jiang, Y., Chen, C., Nguyen, D., Mervak, B.~M., and Tan, C.
\newblock {GPT}-{4V} cannot generate radiology reports yet.
\newblock \emph{ArXiv}, abs/2407.12176, 2024.
\newblock URL \url{https://api.semanticscholar.org/CorpusID:271244474}.

\bibitem[Johnson et~al.(2019)Johnson, Pollard, Berkowitz, Mark, and Horng]{johnson2019mimic-cxr-dataset}
Johnson, A. E.~W., Pollard, T.~J., Berkowitz, S.~J., Mark, R.~G., and Horng, S.
\newblock {MIMIC-CXR} database (version 2.0.0).
\newblock PhysioNet, 2019.

\bibitem[Lad et~al.(2024)Lad, Gurnee, and Tegmark]{lad2024remarkable}
Lad, V., Gurnee, W., and Tegmark, M.
\newblock The remarkable robustness of {LLMs}: Stages of inference?
\newblock \emph{arXiv preprint arXiv:2406.19384}, 2024.

\bibitem[Le et~al.(2024)Le, Patel, Shen, Martin, Eng, Shah, Grullon, and Juyal]{le_learning_2024}
Le, N.~M., Patel, N., Shen, C., Martin, B., Eng, A., Shah, C., Grullon, S., and Juyal, D.
\newblock Learning biologically relevant features in a pathology foundation model using sparse autoencoders.
\newblock In \emph{NeurIPS 2024 Workshop on Advancements In Medical Foundation Models}, December 2024.
\newblock URL \url{https://openreview.net/forum?id=daV16mhUBd}.

\bibitem[Lee et~al.(2006)Lee, Battle, Raina, and Ng]{lee2006efficient}
Lee, H., Battle, A., Raina, R., and Ng, A.
\newblock Efficient sparse coding algorithms.
\newblock \emph{Advances in neural information processing systems}, 19, 2006.

\bibitem[Li et~al.(2023{\natexlab{a}})Li, Patel, Vi{\'e}gas, Pfister, and Wattenberg]{li2023inferencetime}
Li, K., Patel, O., Vi{\'e}gas, F., Pfister, H., and Wattenberg, M.
\newblock Inference-time intervention: Eliciting truthful answers from a language model.
\newblock \emph{Advances in Neural Information Processing Systems}, 36:\penalty0 41451--41530, 2023{\natexlab{a}}.

\bibitem[Li et~al.(2023{\natexlab{b}})Li, Lin, Chen, Lin, Liang, and Chang]{li2023dynamic}
Li, M., Lin, B., Chen, Z., Lin, H., Liang, X., and Chang, X.
\newblock Dynamic graph enhanced contrastive learning for chest x-ray report generation.
\newblock In \emph{Proceedings of the IEEE/CVF Conference on Computer Vision and Pattern Recognition}, pp.\  3334--3343, 2023{\natexlab{b}}.

\bibitem[Lieberum et~al.(2024)Lieberum, Rajamanoharan, Conmy, Smith, Sonnerat, Varma, Kram{\'a}r, Dragan, Shah, and Nanda]{lieberum2024gemmascope}
Lieberum, T., Rajamanoharan, S., Conmy, A., Smith, L., Sonnerat, N., Varma, V., Kram{\'a}r, J., Dragan, A., Shah, R., and Nanda, N.
\newblock {G}emma {S}cope: Open sparse autoencoders everywhere all at once on {G}emma 2.
\newblock \emph{arXiv preprint arXiv:2408.05147}, 2024.

\bibitem[Liu et~al.(2023)Liu, Li, Wu, and Lee]{liu2023llava}
Liu, H., Li, C., Wu, Q., and Lee, Y.~J.
\newblock Visual instruction tuning.
\newblock In \emph{Advances in Neural Information Processing Systems}, volume~36, pp.\  34892--34916, 2023.

\bibitem[Lou et~al.(2025)Lou, Li, Ji, and Yang]{lou_sae-v_2025}
Lou, H., Li, C., Ji, J., and Yang, Y.
\newblock {SAE}-{V}: {Interpreting} {Multimodal} {Models} for {Enhanced} {Alignment}, February 2025.
\newblock URL \url{http://arxiv.org/abs/2502.17514}.
\newblock arXiv:2502.17514 [cs].

\bibitem[Marks et~al.(2024)Marks, Karvonen, and Mueller]{marks2024dictionary_learning}
Marks, S., Karvonen, A., and Mueller, A.
\newblock dictionary\_learning.
\newblock \url{https://github.com/saprmarks/dictionary_learning}, 2024.

\bibitem[Minder et~al.(2025)Minder, Dumas, Juang, Chugtai, and Nanda]{Minder2025RobustlyIC}
Minder, J., Dumas, C., Juang, C., Chugtai, B., and Nanda, N.
\newblock Robustly identifying concepts introduced during chat fine-tuning using crosscoders.
\newblock \emph{arXiv preprint arXiv:2504.02922}, 2025.

\bibitem[Mudide et~al.(2024)Mudide, Engels, Michaud, Tegmark, and de~Witt]{mudide2024efficient}
Mudide, A., Engels, J., Michaud, E.~J., Tegmark, M., and de~Witt, C.~S.
\newblock Efficient dictionary learning with switch sparse autoencoders.
\newblock \emph{arXiv preprint arXiv:2410.08201}, 2024.

\bibitem[O'Brien et~al.(2024)O'Brien, Majercak, Fernandes, Edgar, Chen, Nori, Carignan, Horvitz, and {Poursabzi-Sangde}]{obrien2024steering}
O'Brien, K., Majercak, D., Fernandes, X., Edgar, R., Chen, J., Nori, H., Carignan, D., Horvitz, E., and {Poursabzi-Sangde}, F.
\newblock Steering language model refusal with sparse autoencoders.
\newblock \emph{arXiv preprint arXiv:2411.11296}, 2024.

\bibitem[Pach et~al.(2025)Pach, Karthik, Bouniot, Belongie, and Akata]{pach_sparse_2025}
Pach, M., Karthik, S., Bouniot, Q., Belongie, S., and Akata, Z.
\newblock Sparse {Autoencoders} {Learn} {Monosemantic} {Features} in {Vision}-{Language} {Models}, April 2025.
\newblock URL \url{http://arxiv.org/abs/2504.02821}.
\newblock arXiv:2504.02821 [cs].

\bibitem[Parekh et~al.(2024)Parekh, Khayatan, Shukor, Newson, and Cord]{parekh2024concept}
Parekh, J., Khayatan, P., Shukor, M., Newson, A., and Cord, M.
\newblock A concept-based explainability framework for large multimodal models.
\newblock \emph{Advances in Neural Information Processing Systems}, 37:\penalty0 135783--135818, 2024.

\bibitem[Park et~al.(2024)Park, Choe, and Veitch]{park2024linear}
Park, K., Choe, Y.~J., and Veitch, V.
\newblock The linear representation hypothesis and the geometry of large language models.
\newblock In \emph{Proceedings of the 41st International Conference on Machine Learning}, pp.\  39643--39666. PMLR, July 2024.
\newblock URL \url{https://proceedings.mlr.press/v235/park24c.html}.

\bibitem[Paulo et~al.(2024)Paulo, Mallen, Juang, and Belrose]{paulo_automatically_2024}
Paulo, G., Mallen, A., Juang, C., and Belrose, N.
\newblock Automatically {Interpreting} {Millions} of {Features} in {Large} {Language} {Models}, December 2024.
\newblock URL \url{http://arxiv.org/abs/2410.13928}.
\newblock arXiv:2410.13928 [cs].

\bibitem[P{\'e}rez-Garc{\'\i}a et~al.(2025)P{\'e}rez-Garc{\'\i}a, Sharma, Bond-Taylor, Bouzid, Salvatelli, Ilse, Bannur, Castro, Schwaighofer, Lungren, et~al.]{perezgarcia2025raddino}
P{\'e}rez-Garc{\'\i}a, F., Sharma, H., Bond-Taylor, S., Bouzid, K., Salvatelli, V., Ilse, M., Bannur, S., Castro, D.~C., Schwaighofer, A., Lungren, M.~P., et~al.
\newblock Exploring scalable medical image encoders beyond text supervision.
\newblock \emph{Nature Machine Intelligence}, 7:\penalty0 119--130, 2025.
\newblock \doi{10.1038/s42256-024-00965-w}.

\bibitem[Quinn et~al.(2021)Quinn, Senadeera, Jacobs, Coghlan, and Le]{quinn2021trust}
Quinn, T.~P., Senadeera, M., Jacobs, S., Coghlan, S., and Le, V.
\newblock Trust and medical ai: the challenges we face and the expertise needed to overcome them.
\newblock \emph{Journal of the American Medical Informatics Association}, 28\penalty0 (4):\penalty0 890--894, 2021.

\bibitem[Shaham et~al.(2024)Shaham, Schwettmann, Wang, Rajaram, Hernandez, Andreas, and Torralba]{shaham2024multimodal}
Shaham, T.~R., Schwettmann, S., Wang, F., Rajaram, A., Hernandez, E., Andreas, J., and Torralba, A.
\newblock A multimodal automated interpretability agent.
\newblock In \emph{Forty-first International Conference on Machine Learning}, 2024.

\bibitem[Simon \& Zou(2024)Simon and Zou]{simon_interplm_2024}
Simon, E. and Zou, J.
\newblock {InterPLM}: {Discovering} {Interpretable} {Features} in {Protein} {Language} {Models} via {Sparse} {Autoencoders}, November 2024.
\newblock URL \url{https://www.biorxiv.org/content/10.1101/2024.11.14.623630v1}.
\newblock Pages: 2024.11.14.623630 Section: New Results.

\bibitem[Stevens et~al.(2025)Stevens, Chao, Berger-Wolf, and Su]{stevens2025sparse}
Stevens, S., Chao, W.-L., Berger-Wolf, T., and Su, Y.
\newblock Sparse autoencoders for scientifically rigorous interpretation of vision models.
\newblock \emph{arXiv preprint arXiv:2502.06755}, 2025.

\bibitem[Templeton et~al.(2024)Templeton, Conerly, Marcus, Lindsey, Bricken, Chen, Pearce, Citro, Ameisen, Jones, Cunningham, Turner, McDougall, MacDiarmid, Freeman, Sumers, Rees, Batson, Jermyn, Carter, Olah, and Henighan]{templeton2024scaling}
Templeton, A., Conerly, T., Marcus, J., Lindsey, J., Bricken, T., Chen, B., Pearce, A., Citro, C., Ameisen, E., Jones, A., Cunningham, H., Turner, N.~L., McDougall, C., MacDiarmid, M., Freeman, C.~D., Sumers, T.~R., Rees, E., Batson, J., Jermyn, A., Carter, S., Olah, C., and Henighan, T.
\newblock Scaling monosemanticity: Extracting interpretable features from {Claude 3 Sonnet}.
\newblock \emph{Transformer Circuits Thread}, 2024.
\newblock URL \url{https://transformer-circuits.pub/2024/scaling-monosemanticity/index.html}.

\bibitem[Tu et~al.(2024)Tu, Azizi, Driess, Schaekermann, Amin, Chang, Carroll, Lau, Tanno, Ktena, Palepu, Mustafa, Chowdhery, Liu, Kornblith, Fleet, Mansfield, Prakash, Wong, Virmani, et~al.]{tu2024medpalmm}
Tu, T., Azizi, S., Driess, D., Schaekermann, M., Amin, M., Chang, P.-C., Carroll, A., Lau, C., Tanno, R., Ktena, I., Palepu, A., Mustafa, B., Chowdhery, A., Liu, Y., Kornblith, S., Fleet, D., Mansfield, P., Prakash, S., Wong, R., Virmani, S., et~al.
\newblock Towards generalist biomedical {AI}.
\newblock \emph{NEJM AI}, 1\penalty0 (3):\penalty0 AIoa2300138, February 2024.
\newblock \doi{10.1056/AIoa2300138}.

\bibitem[Turner et~al.(2023)Turner, Thiergart, Leech, Udell, Vazquez, Mini, and MacDiarmid]{turner2023steering}
Turner, A.~M., Thiergart, L., Leech, G., Udell, D., Vazquez, J.~J., Mini, U., and MacDiarmid, M.
\newblock Steering language models with activation engineering.
\newblock \emph{arXiv preprint arXiv:2308.10248}, 2023.

\bibitem[Wang et~al.(2022)Wang, Variengien, Conmy, Shlegeris, and Steinhardt]{wang2022interpretability}
Wang, K., Variengien, A., Conmy, A., Shlegeris, B., and Steinhardt, J.
\newblock Interpretability in the wild: a circuit for indirect object identification in {GPT-2} small.
\newblock \emph{arXiv preprint arXiv:2211.00593}, 2022.

\bibitem[Wang et~al.(2023)Wang, Liu, Wang, and Zhou]{wang2023metransformer}
Wang, Z., Liu, L., Wang, L., and Zhou, L.
\newblock Metransformer: Radiology report generation by transformer with multiple learnable expert tokens.
\newblock In \emph{Proceedings of the IEEE/CVF Conference on Computer Vision and Pattern Recognition}, pp.\  11558--11567, 2023.

\bibitem[Wattenberg \& Viégas(2024)Wattenberg and Viégas]{wattenberg2024relationalcompositionneuralnetworks}
Wattenberg, M. and Viégas, F.~B.
\newblock Relational composition in neural networks: A survey and call to action, 2024.
\newblock URL \url{https://arxiv.org/abs/2407.14662}.

\bibitem[Wu et~al.(2025)Wu, Arora, Geiger, Wang, Huang, Jurafsky, Manning, and Potts]{wu_axbench_2025}
Wu, Z., Arora, A., Geiger, A., Wang, Z., Huang, J., Jurafsky, D., Manning, C.~D., and Potts, C.
\newblock {AxBench}: {Steering} {LLMs}? {Even} {Simple} {Baselines} {Outperform} {Sparse} {Autoencoders}, March 2025.
\newblock URL \url{http://arxiv.org/abs/2501.17148}.
\newblock arXiv:2501.17148 [cs].

\bibitem[Yan et~al.(2024)Yan, He, Yue, and Wang]{Yan2024WorseTR}
Yan, Q., He, X., Yue, X., and Wang, X.~E.
\newblock Worse than random? an embarrassingly simple probing evaluation of large multimodal models in medical {VQA}.
\newblock \emph{ArXiv}, abs/2405.20421, 2024.
\newblock URL \url{https://api.semanticscholar.org/CorpusID:270199350}.

\bibitem[Yang et~al.(2024)Yang, Xu, Sellergren, Kohlberger, Zhou, Ktena, Kiraly, Ahmed, Hormozdiari, Jaroensri, et~al.]{yang2024advancing}
Yang, L., Xu, S., Sellergren, A., Kohlberger, T., Zhou, Y., Ktena, I., Kiraly, A., Ahmed, F., Hormozdiari, F., Jaroensri, T., et~al.
\newblock Advancing multimodal medical capabilities of {G}emini.
\newblock \emph{arXiv preprint arXiv:2405.03162}, 2024.

\bibitem[Yildirim et~al.(2024)Yildirim, Richardson, Wetscherek, Bajwa, Jacob, Pinnock, Harris, Coelho De~Castro, Bannur, Hyland, et~al.]{yildirim2024multimodal}
Yildirim, N., Richardson, H., Wetscherek, M.~T., Bajwa, J., Jacob, J., Pinnock, M.~A., Harris, S., Coelho De~Castro, D., Bannur, S., Hyland, S., et~al.
\newblock Multimodal healthcare {AI}: Identifying and designing clinically relevant vision-language applications for radiology.
\newblock In \emph{Proceedings of the 2024 CHI Conference on Human Factors in Computing Systems}, pp.\  1--22, 2024.

\bibitem[Zhang et~al.(2024)Zhang, Shen, Li, and Liu]{zhang_large_2024}
Zhang, K., Shen, Y., Li, B., and Liu, Z.
\newblock Large {Multi}-modal {Models} {Can} {Interpret} {Features} in {Large} {Multi}-modal {Models}, November 2024.
\newblock URL \url{http://arxiv.org/abs/2411.14982}.
\newblock arXiv:2411.14982 [cs].

\bibitem[Zhou et~al.(2024)Zhou, Adithan, Acosta, Topol, and Rajpurkar]{zhou2024generalist}
Zhou, H.-Y., Adithan, S., Acosta, J.~N., Topol, E.~J., and Rajpurkar, P.
\newblock A generalist learner for multifaceted medical image interpretation.
\newblock \emph{arXiv preprint arXiv:2405.07988}, 2024.

\end{thebibliography}
\bibliographystyle{icml2025}

\newpage
\appendix
\onecolumn

\numberwithin{figure}{section}
\numberwithin{table}{section}

\section{Additional experimental setup details}

\subsection{Representation filtering}\label{sec:app:act-fil}
\cref{tab:act-fil} shows the effect of the token-based representation filtering, keeping only the most relevant tokens for the task at hand. We discard any fixed segments from the original prompt template: the system prompt, beginning- and end-of-sequence tokens, the chat template delimiters, and the fixed parts of the instruction. We choose to depict an example that include all three possible \ac{CXR} image views as a comprehensive representation of the filtering.
\begin{table*}[h]
\centering
\caption{Example of token based representations filtering, including a frontal, lateral and prior image. In the filtered prompt, we indicate the indices of the tokens which were kept in square brackets. We further extract the message type (`human' or `assistant') and the content type (`str' or `image').}
\label{tab:act-fil}
\begin{tabular}{p{8cm}p{8cm}}
\toprule
Original Prompt & Filtered Prompt \\
\midrule
\begin{Verbatim}[fontsize=\scriptsize]
<s>, _You, _are, _an, _expert, _radi, ology, _assistant,
_task, ed, _with, _interpre, ting, _a, _ch, est, 
_X, -, ray, _study, ., 
_, _US, ER, :, _, 
_Given, _the, _current, _front, al, _image, <image>x1369, 
_the, _current, _later, al, _image, <lat_image>x1369, 
_and, _the, _prior, _front, al, _image, <prev_im>x1369,
_P, RI, OR, _, RE, PORT, :, _N, /, A,
_Prov, ide, _a, _description, _of, _the, _find, ings, _in, 
_the, _radi, ology, _study, _in, _comparison, _to, _the,
_prior, _front, al, _image,., 
_IN, D, ICATION, :, __, _year, _old, _woman, _with, 
_rec, urrent, _asp, iration, _p, na, ,, _now, _with, 
_f, lare, _in, _s, put, um, ,, _c, ough, ,, _and, 
_bil, ater, al, _lower, _lo, be, _crack, les, _//, 
_assess, _for, _new, _p, neum, onia,
_TE, CH, NI, QUE, :, _Ch, est, _radi, ograph, ,,
_PA, _and, _later, al, _views,
_CO, MP, AR, I, SON, :, _Thus, _radi, ograph, __, _,
_A, SS, IST, ANT, :, 
_Bil, ater, al, _lower, _lo, be, _op, ac, ities, _are, 
_improved, _compared, _to, __, ., _There, _are, _small, 
_co, ales, c, ence, _into, _several, _nod, ular, 
_op, ac, ities, _remaining, _on, _the, _right, _but, 
_mostly, _improved, ., _L, ungs, _are, _m, ild, ly,
_hyper, infl, ated, ., _There, _is, _no, _definite, 
_ple, ural, _eff, usion, ., _Card, iom, ed, iast, inal,
_sil, hou, ette, _is, _normal, _size, ., _,2, _f, ract, ured,
_sc, rew, s, _in, _right, _hum, eral, _head, _is, _un, 
changed, _from, _prior, ., </s>
\end{Verbatim}
&
\begin{Verbatim}[fontsize=\scriptsize]
[start_index:end_index] message_type content_type [filtered tokens]

[31:32] human str ['_image']
[1400:1401] human image ['<image>']
[1405:1406] human str ['_image']
[2774:2775] human image ['<lat_image>']
[2780:2781] human str ['_image']
[4149:4150] human image ['<prev_im>']
[4150:4160] human str ['_P', 'RI', 'OR', '_', 
'RE', 'PORT', ':', '_N', '/', 'A']
[4182:4226] human str ['_IN', 'D', 'ICATION', ':', '__', 
            '_year', '_old', '_woman', '_with', '_rec', 
            'urrent', '_asp', 'iration', '_p', 'na', ',', 
            '_now', '_with', '_f', 'lare', '_in', '_s', 
            'put', 'um', ',', '_c', 'ough', ',', '_and',
            '_bil', 'ater', 'al', '_lower', '_lo', 'be',
            '_crack', 'les', '_//', '_assess', '_for',
            '_new', '_p', 'neum', 'onia']
[4226:4241] human str ['_TE', 'CH', 'NI', 'QUE', ':',
            '_Ch', 'est', '_radi', 'ograph', ',', 
            '_PA', '_and', '_later', 'al', '_views']
[4241:4251] human str ['_CO', 'MP', 'AR', 'I', 'SON', ':', 
            '_Thus', '_radi', 'ograph', '__']
[4257:4344] assistant str ['_Bil', 'ater', 'al', '_lower',
            '_lo', 'be', '_op', 'ac', 'ities', '_are', 
            '_improved', '_compared', '_to', '__', '.', 
            '_There', '_are', '_small', '_co', 'ales', 
            'c', 'ence', '_into', '_several', '_nod', 'ular', 
            '_op', 'ac', 'ities', '_remaining', '_on', '_the',
            '_right', '_but', '_mostly', '_improved', '.', 
            '_L', 'ungs', '_are', '_m', 'ild', 'ly', 
            '_hyper', 'infl', 'ated', '.', '_There', '_is',
            '_no', '_definite', '_ple', 'ural', '_eff', 'usion',
            '.',  '_Card', 'iom', 'ed', 'iast', 'inal', '_sil',
            'hou', 'ette', '_is', '_normal', '_size', '.', '_', 
            '2', '_f', 'ract', 'ured', '_sc', 'rew', 's', '_in', 
            '_right',  '_hum', 'eral', '_head', '_is', '_un',
            'changed', '_from', '_prior', '.']
\end{Verbatim}
\\
\midrule
Total \# tokens = 4348 & Total \# filtered tokens = 176 \\
\bottomrule
\end{tabular}
\end{table*}

\clearpage
\subsection{SAE Hyperparameters}\label{sec:app:sae-hp}
\cref{tab:sae-hparams} shows all the hyperparameters that were overridden from the default values in the \texttt{dictionary\_learning} repository\footnote{\url{https://github.com/saprmarks/dictionary_learning}}. 

We realised we had accidentally run with reversed Matryoshka group sizes, with the largest group being `innermost'. This appears to have slightly reduced the overall interpretability of the discovered features. We will quantify the impact on steering in a revision. However, given we do not see a strong association between interpretability score and steering success on the feature level, we do not expect a significant difference in our findings.
\begin{table*}[h]
\centering
\caption{Hyperparameters used for training the \ac{SAE} using the open-source repository \texttt{dictionary\_learning}.}
\label{tab:sae-hparams}
\footnotesize
\begin{tabular}{@{}llp{9cm}@{}}
\toprule
\textbf{Category} & \textbf{Value} & \textbf{Description} \\
\midrule
\multicolumn{3}{@{}l}{\textit{SAE Configuration}} \\
\quad Type & \texttt{matryoshka\_batch\_topk} & Type of \ac{SAE} architecture used. \\
\quad Activation dimension & 4096 & The dimensionality of the activations in the \ac{SAE} model. \\
\quad Expansion factor & 8 & The factor by which the number of dictionary atoms is expanded. \\
\quad Layer ID & 15 & The specific layer from which the embeddings are extracted. \\
\quad Hookpoint & \texttt{residual} & The hookpoint in the model from where embeddings are captured. \\
\quad $k$ (active features) & 256 & The number of active features in the sparse encoding. \\
\quad Group fractions & [1/2, 1/4, 1/8, 1/16, 1/16] & The fractions used to group dictionary atoms, controlling the sparsity levels at different scales. \\
\midrule
\multicolumn{3}{@{}l}{\textit{Trainer Configuration}} \\
\quad Epochs & 1 & The number of training epochs. \\
\quad Learning rate & Auto & The learning rate is automatically set by the trainer. \\
\quad Aux $k$ alpha & 0.03125 & The weight of the auxiliary $k$-sparse loss term. \\
\quad Threshold beta & 0.999 & The exponential decay rate used for thresholding. \\
\quad Threshold start step & 1000 & The step at which thresholding begins during training. \\
\quad Threshold dead features & 100000 & The number of tokens after which a feature without activation will be considered ``dead'' (used in the auxiliary loss). \\
\midrule
\multicolumn{3}{@{}l}{\textit{Data Configuration}} \\
\quad Batch size & 8192 & The number of samples in each batch for training. \\
\quad Normalize activations & True & Whether or not to normalize the activations to unit norm. \\
\bottomrule
\end{tabular}
\end{table*}

\subsection{Auxiliary Loss}\label{sec:app:aux-loss}

As introduced by \citet{gao_scaling_2024}, the auxiliary loss $\mathcal{L}_{\mathrm{aux}}$ captures the reconstruction error associated with the top-$k_{aux}$ dead nodes in the sparse latent space. It is defined as:

\begin{equation}
    \mathcal{L}_{\mathrm{aux}} = \left\| e - \hat{e} \right\|_2^2,
\end{equation}

where $e = x - \hat{x}$ is the error from the main model, and $\hat{e} = W_{dec}^kz$ represents the reconstruction obtained using the top-$k_{aux}$ dead latents. A latent is considered "dead" if it has not activated for a predefined number of tokens, as specified by the threshold dead features.

\clearpage
\FloatBarrier
\section{Feature statistics}
\begin{figure}[h]
    \includegraphics[width=\textwidth]{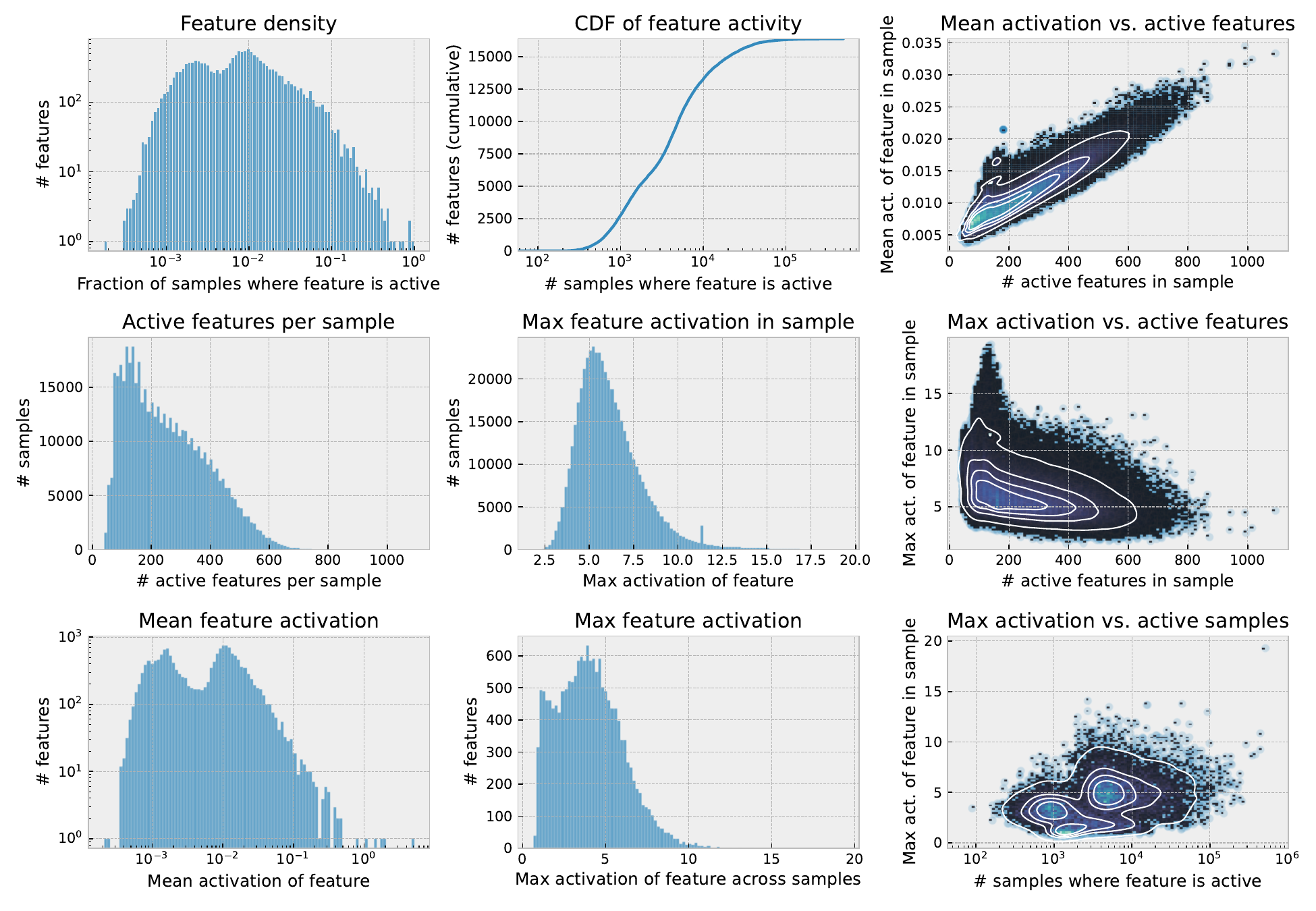}
    \caption{Statistics on the activation frequency and strength of features in the studied SAE, using a random sample of 500,000 data points from the training set. Each sample here refers to a single token position.}
    \label{fig:basic_latent_stats}
\end{figure}

\Cref{fig:basic_latent_stats} provides statistics on the activation frequency of features in the studied \ac{SAE}, using a random sample of 500,000 data points from the training set. Each sample here refers to the model representation extracted at a single token position.

Defining the feature density as the fraction of samples on which a given feature activates, we observe (\Cref{fig:basic_latent_stats}; top left, top middle) a range of values, with a very small number of features activating on almost all samples. Unfortunately, these commonly-occurring features are not well-explained: \feature{7633} activates on 99.3\% of samples and achieves $\fone=0.40$, \feature{6727} activates on 86.8\% of samples and is described as \featexpl{Detailed comparison with prior images highlighting interval changes.} with $\fone=0.67$. We did not observe a correlation between feature density and interpretability (detection \fone), however some of the most interpretable features have low density - \feature{13515} ($\fone=0.97$, \featexpl{Elevation of the hemidiaphraghm.}) activates on only 159 samples ($0.03\%$).

\FloatBarrier
\section{Automated interpretability}\label{sec:app:automated_interpretability}
We split the dataset 50:50 by patient into `train' and `evaluation' for the purpose of generating and scoring latent interpretations respectively. We consider each latent individually.
We use GPT-4o (2024-11-20) via the Azure OpenAI service. Initial experiments with o3-mini (2025-01-31) yielded inferior results, but may benefit from further prompt engineering.

\subsection{Selecting and preparing feature exemplars}
For each feature, we select `exemplar' data samples by selecting an equal number of non-activating and activating samples, for a total of 50. Activating samples are selected from the top decile of the feature activation distribution, treating an activation of 0 as the bottom decile. Experiments with stratified sampling indicated this strategy was marginally superior.

Each data sample refers to a specific token index for a specific input sequence. To explain the feature activation at that point, we provide the \ac{LLM} with the full sequence up to that token, and the next 100 characters. We indicate the `current' token with double square brackets. We found that truncating to 100 characters after the current token provided slightly improved interpretability scores relative to truncating to 0, 10, or 250 characters. Note that our few-shot examples, being hand-designed, inconsistently adhere to the truncation length.

\begin{figure}
    \centering
    \includegraphics[width=0.5\textwidth]{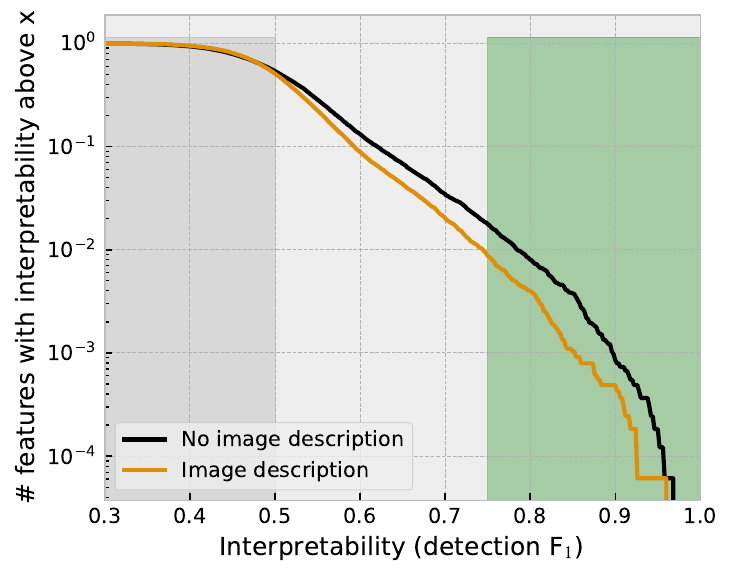}
    \caption{Including the `description' of the image (`Findings' section of the report) to the interpretation \ac{LLM} does not increase the number of interpretable features we discover.}
    \label{fig:use_of_image}
\end{figure}

Since \mairatwo is a \ac{MLLM}, it receives interleaved images and text as input. Our interpretation \ac{LLM} is text-only, we replace the image tokens with the string `\verb|<image>|'. In a chest X-ray report, the `Findings' section of the report approximates a description of the image. We experimented with including this section prefixed to the data sample, to provide the \ac{LLM} with information about the image contents. Surprisingly (\Cref{fig:use_of_image}), we found this did \emph{not} improve the number of interpretable features we could discover. We note however that we train the SAE (and hence interpret it) on both the prompt and target tokens used for \mairatwo training, where the training target is the `Findings' section. Hence, in many cases the input text to the model may already contain some part of the image description.

\subsection{Generating interpretations}
\label{sec:app:auto_interp}
We use a modified prompt based on \citet{paulo_automatically_2024}\footnote{ \url{https://github.com/EleutherAI/delphi/tree/article_version}}.
For each example, we present a single latent activation, scaled to 0-9 based on the range of observed values of that latent, following \citet{gao_scaling_2024}.

The instruction provided to the explanation-deriving LLM is shown below:

\begin{promptbox}[Latent interpretation: system message]
You are a meticulous AI researcher conducting an important investigation into the activation patterns of a large autoregressive vision-language model trained on chest X-ray data. You will be presented with samples of prompts to and outputs from this model with corresponding activation levels at a specified token. You task is to analyze this data and provide an explanation which succinctly encapsulates patterns to explain the observed activation levels.\\
\\
Guidelines:\\
- Each data example consists of some preamble text, the [[current token]], and the next few tokens, as well as an "activation level" computed at the [[current token]]. Note that the current token is delimited with "[[, ]]".\\
- The activation level indicates how representative the sample is of the pattern we wish to understand.\\
- Activation levels close to zero mean the pattern is NOT present.\\
- These examples can refer to "<image>". In this case, an <IMAGE\_DESCRIPTION> may be provided.\\
\\
Produce the SHORTEST and MOST CONCISE explanation of the pattern, with a rationale.\\
\\
Respond in JSON with the following fields:
\begin{verbatim}
{
    "rationale": "Justification for this explanation.",
    "explanation": "Concise explanation of the pattern."
}
\end{verbatim}
\end{promptbox}

We use six few-shot examples based on our medical use-case, devised in conjunction with Claude 3.7 Sonnet Thinking. Each example features samples from a hypothetical latent with different activation levels. We tried to design few-shots to cover possible latents including those pertaining to pathologies or other image-specific concepts, as well as `meta' concepts such as the length into the input sequence (as described by \citet{gao_scaling_2024}), and `superficial' concepts such as the appearance of a specific token.
 
 We present one below:
 \begin{promptbox}[Latent interpretation: illustrative few-shot example]
 Input:\\
 \\
 Example 1: [[\_IN]]DICATION: 58-year-old female with persistent cough. FINDINGS: No acute cardiopulmonary process. Heart size is normal.\\
 Activation: 8\\
 Example 2: TECHNIQUE: PA and lateral chest radiographs were obtained. [[The]] cardiac silhouette appears normal in size.\\
 Activation: 0\\
 Example 3: EXAMINATION: Chest CT. [[\_IN]]DICATION: Follow-up for previously identified pulmonary nodule. FINDINGS: Left lung is clear, nodule persists in right lower lobe. No new masses identified.\\
 Activation: 7\\
 Example 4: The patient presents with shortness of breath and chest pa[[\_in]]. No fever reported.\\
 Activation: 0\\
 ...\\
 \\
 Output:
 \begin{verbatim}
{
    "rationale": "The activation is high when '_IN' appears as the first token of the 'INDICATION' section, 
    but is not high when '_IN' appears in other contexts.",
    "explanation": "The token '_IN' appearing as part of the word 'INDICATION'"
}
 \end{verbatim}
 \end{promptbox}

\subsection{Scoring interpretations}
\label{sec:app:interp_scoring}
We follow the detection scoring approach outlined by \citet{paulo_automatically_2024}, and again draw inspiration from the prompts used in that study. The LLM is provided with the interpretation of a latent (a sentence) and a sample. It is asked to classify whether the latent would activate on that sample.

Below we show the system message:
\begin{promptbox}[Interpretation scoring: system message]
You are an intelligent and meticulous researcher into clinical language, with a specialisation in radiology.\\
\\
You will be provided with a "latent explanation", which describes a "latent property" of the text (a concept or pattern which can appear in text), such as "patient deterioration" or "mention of lung opacities".\\
\\
You will then be given several text examples. Your task is to determine which examples possess latent property. These examples may appear to have truncated text. Regardless of the formatting, focus on determining whether the text has the property in the "latent explanation".\\
\\
For each example, return 1 if the example demonstrates the property, and 0 otherwise.
\end{promptbox}

Again we devised few-shot examples in conjunction with Claude 3.7 Sonnet Thinking.  Using a small development dataset, we found that the LLM performed better on the task when it was asked to classify a single sample at a time, however few-shot examples included several samples at once. An illustrative case is shown below:
\begin{promptbox}[Interpretation scoring: illustrative few-shot example]
Input:\\
\\
Latent explanation: Comparison with prior imaging studies.\\
\\
Test examples:\\
\\
Example 0: Chest radiograph demonstrates bibasilar atelectasis with small bilateral pleural effusions. Heart size is upper limits of normal.\\
Example 1: Interval development of small right pleural effusion not present on prior chest radiograph performed 2 days ago.\\
\\
Output:\\
\\
{[0, 1]}
\end{promptbox}

\section{Automated steering evaluation}\label{sec:app:steering_evaluation}
Our automated evaluation of steering success is implemented using a LLM judge similar to \citet{wu_axbench_2025}. The model is prompted to provide a score from 0 to 1 for on-target changes and off-target changes and to reason about its decision beforehand. The LLM judge is then provided with ten few-shot examples of the task. We use individual system messages and few-shot examples for positive and negative steering.
Similar to the automated interpretability pipeline (c.f. \cref{sec:app:automated_interpretability}), we use GPT-4o (2024-11-20) via the Azure OpenAI service.

Below we show the system message.
\begin{promptbox}[Automated Steering Evaluation: system message]
You are an intelligent and meticulous researcher into clinical language, with a specialisation in radiology.\par
\vspace{1em}
You will be provided with a "concept", which describes a concept or pattern which can appear in text, such as "patient deterioration" or "mention of lung opacities".\par
\vspace{1em}
You will further be provided with two text samples:\par
\vspace{1em}
1. A findings section of a radiology report (called "original")\\
2. A modified version of that section (called "modified")\\
\\
You will then be given several text examples. Please act as an impartial judge and determine (1) to which degree the modified version is changed to \textbf{\{\{MODIFIER\}\}} better represent the concept compared to the original version ("on\_target\_score") and (2) to which degree the rest of the report stayed unchanged compared to the original version ("off\_target\_score").\par
\vspace{1em}
Focus solely on the changes in information, without regard for different ways of representing the same information (e.g. reordering, different wording, etc.). Changes in relation to the concept should NOT be counted to the "off\_target\_score". Report both scores on a scale from 0 to 1 where 0 means no change and 1 means strong changes.\par
\vspace{1em}
Respond in JSON with the following fields:
\begin{verbatim}
{
    "on_target_score_reasoning": <reasoning for on_target_score>,
    "off_target_score_reasoning": <reasoning for off_target_score>,
    "on_target_score": <score between 0 and 1>,
    "off_target_score": <score between 0 and 1>
}
\end{verbatim}
\end{promptbox}

The value of \textbf{\texttt{\{\{MODIFIER\}\}}} in the system message depends on the steering direction and equals to \say{better represents} and \say{SUPPRESS} for steering in the positive and negative direction, respectively.

To calibrate the scores, few-shot examples are provided. An illustrative case is shown below with on-target changes between the original and the modified report highlighted in \textcolor{ForestGreen}{green}, and off-target changes highlighted in \textcolor{Orange}{orange}.

\begin{promptbox}[Automated Steering Evaluation: illustrative few-shot example]
Original report: Cardiac size is within normal limits. The lungs are clear \textcolor{Orange}{without focal consolidation}, pleural effusion, or pneumothorax.\par
\vspace{1em}  %
Modified report: Cardiac size is normal. \textcolor{ForestGreen}{The aorta is tortuous.} The lungs are clear. There is no pneumothorax or pleural effusion.\par
\vspace{1em}
Concept: Increased tortuosity or calcification of the thoracic aorta.\par
\vspace{1em}
Output:\\
\begin{verbatim}
{
    "on_target_score_reasoning": "The modified report contains the concept in the statement \"The aorta is
        tortuous.\". However, it doesn't mention an increase.",
    "off_target_score_reasoning": "The explicit mention of the absence of focal consolidation is omitted
        in the modified report.",
    "on_target_score": 0.7,
    "off_target_score": 0.2
}
\end{verbatim}
\end{promptbox}

\section{Feature steering}\label{app:feature_steering_details}

In this section, we provide further results on feature steering and outline the selection process of features for these experiments.

\subsection{Selection of features for steering}

Since steering and its evaluation is computationally expensive, we select only a subset of features for this experiment. Since steering evaluation relies on a meaningful description of a feature, we start with an initial selection of features with $\fone > 0.85$. Based on early results that indicate higher steering success of features that activate frequently in the training data, we manually add more such features, again focussing on ones with high \fone-scores. These 74 features are depicted in \Cref{tab:features}.
Our objective here was to find the best-case scenario for steering to understand whether it is possible, hence iteratively searching for steerable features.

\begin{figure*}
    \centering
    \includegraphics[width=0.92\textwidth]{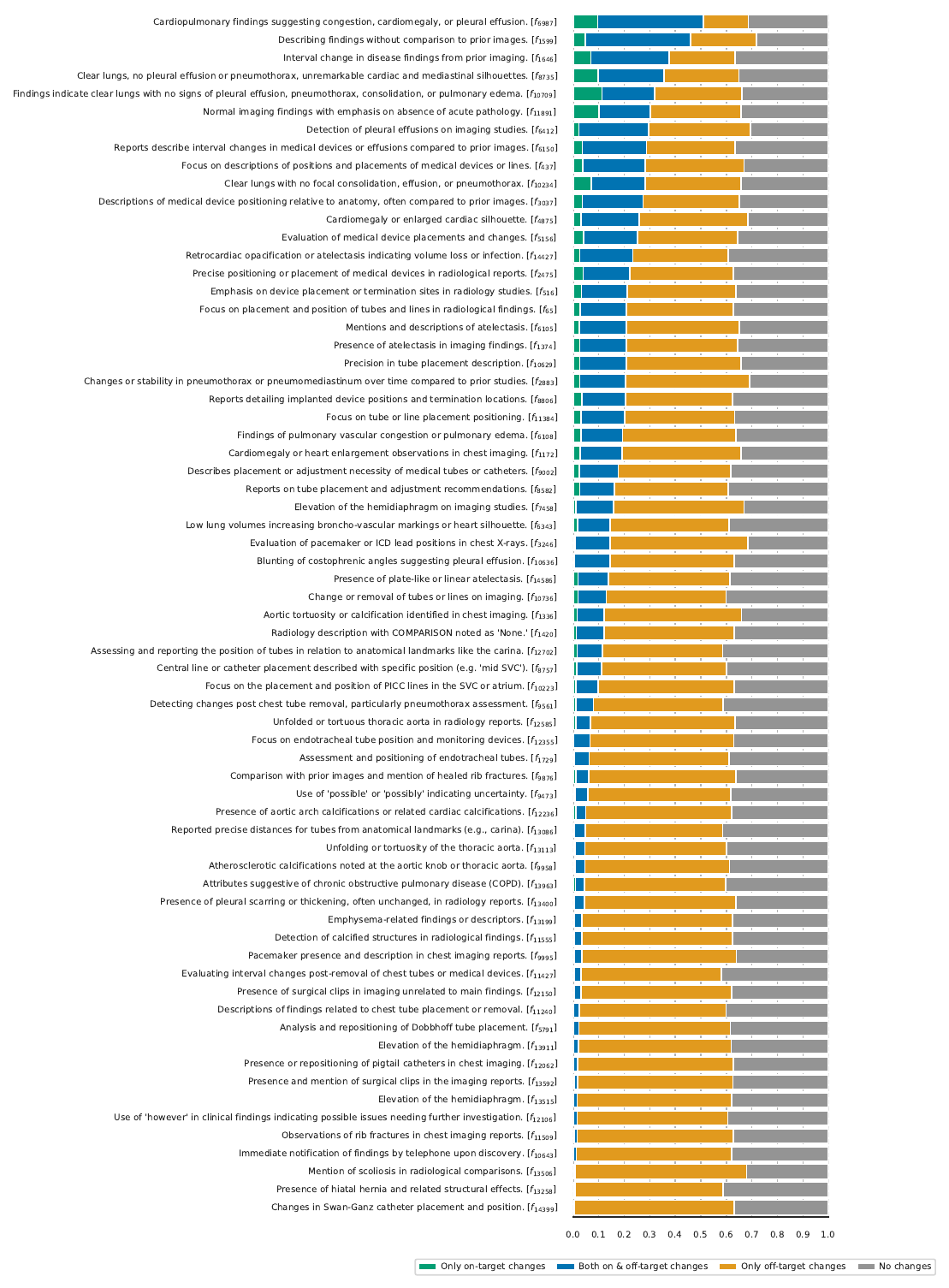}
    \caption{Evaluating success of steering with $\alpha=10$ for all steered features. Proportions of pure \textit{on-target changes}, \textit{off-target changes}, the combination of the two and no changes at all. The results of the automatic steering evaluation are binarised for this plot. Scores $>0.1$ count as changes.}
    \label{fig:pos-steering-all}
\end{figure*}

\begin{figure*}
    \centering
    \includegraphics[width=0.92\textwidth]{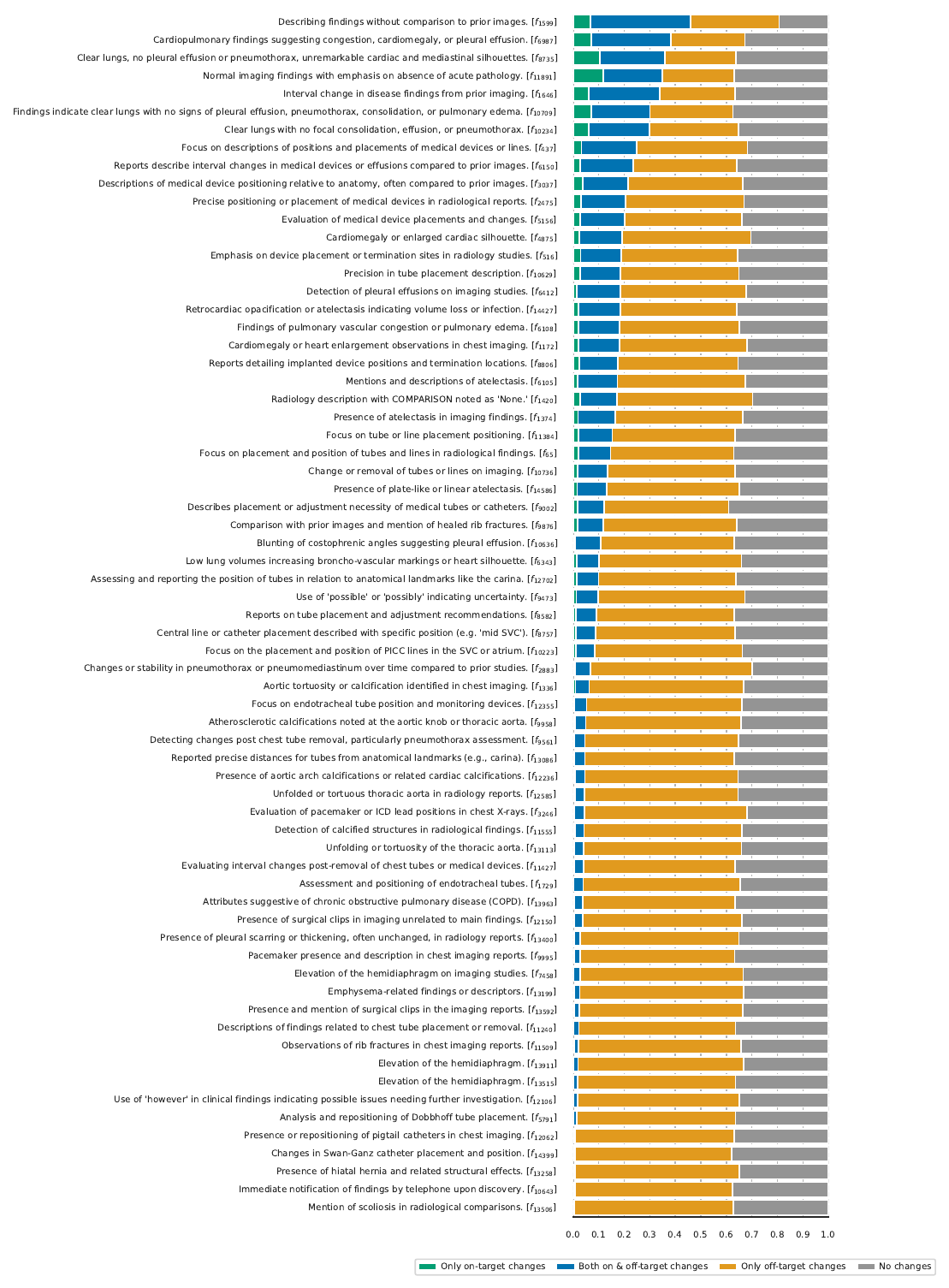}
    \caption{Evaluating success of steering with $\alpha=-10$ for all steered features. Proportions of pure \textit{on-target changes}, \textit{off-target changes}, the combination of the two and no changes at all. The results of the automatic steering evaluation are binarised for this plot. Scores $>0.1$ count as changes.}
    \label{fig:neg-steering-all}
\end{figure*}

To obtain the final set of features for the steering task, we then remove findings that are not `output' features \citep{paulo_automatically_2024} (e.g. \feature{1420} \featexpl{Radiology description with COMPARISON noted as 'None'}), and features describing low lung volumes. The steering evaluation results for the resulting set of 67 features are shown in \Cref{fig:pos-steering-all} with corresponding examples in \Cref{tab:more-steering-examples} for positive steering and \Cref{fig:neg-steering-all} and \Cref{tab:neg-steering-examples} for negative steering.

For the analysis of the correlation between the activation frequency and on-target / off-target scores in \Cref{sec:findings_steering}, we did not include the manually added highly activating features to avoid further bias than the one induced by restricting the steering analyses to highly interpretable features ($\fone > 0.85$). This resulted in 50 samples for this analysis. Due to the low sample size, we used a permutation test with 9,999 permutations to check for statistical significance.

\subsection{Detailed feature steering results}
\begingroup
\centering
\footnotesize
\begin{longtable}{@{} >{\raggedright\arraybackslash} p{12mm} lcr p{11cm} @{}}
\caption{Feature explanations, their validation \fone scores, and number of active samples, grouped by common themes.}
\label{tab:features} \\
\toprule
\textbf{Group} & \textbf{ID} & $\mathbf{\fone}$ & \textbf{\# Active} & \textbf{Explanation} \\
\midrule
\endfirsthead
\endfoot
\caption{Feature explanations, their validation \fone scores, and number of active samples, grouped by common themes (cont.).} \\
\midrule
\textbf{Group} & \textbf{ID} & $\mathbf{\fone}$ & \textbf{\# Active} & \textbf{Explanation} \\
\midrule
\endhead
\multirow{2}{=}{Medical devices} & \feature{8806} & 0.86 & 1609 & Reports detailing implanted device positions and termination locations. \\
 & \feature{516} & 0.85 & 10639 & Emphasis on device placement or termination sites in radiology studies. \\
 & \feature{9995} & 0.85 & 831 & Pacemaker presence and description in chest imaging reports. \\
 & \feature{12062} & 0.95 & 1028 & Presence or repositioning of pigtail catheters in chest imaging. \\
 & \feature{11240} & 0.94 & 1365 & Descriptions of findings related to chest tube placement or removal. \\
 & \feature{10736} & 0.86 & 1059 & Change or removal of tubes or lines on imaging. \\
 & \feature{11427} & 0.86 & 3109 & Evaluating interval changes post-removal of chest tubes or medical devices. \\
 & \feature{9561} & 0.86 & 2090 & Detecting changes post chest tube removal, particularly pneumothorax assessment. \\
 & \feature{13086} & 0.90 & 665 & Reported precise distances for tubes from anatomical landmarks (e.g., carina). \\
 & \feature{10223} & 0.88 & 2908 & Focus on the placement and position of PICC lines in the SVC or atrium. \\
 & \feature{8582} & 0.87 & 762 & Reports on tube placement and adjustment recommendations. \\
 & \feature{12702} & 0.86 & 329 & Assessing and reporting the position of tubes in relation to anatomical landmarks like the carina. \\
 & \feature{437} & 0.86 & 24497 & Focus on descriptions of positions and placements of medical devices or lines. \\
 & \feature{10629} & 0.86 & 1947 & Precision in tube placement description. \\
 & \feature{11384} & 0.86 & 611 & Focus on tube or line placement positioning. \\
 & \feature{9002} & 0.85 & 5615 & Describes placement or adjustment necessity of medical tubes or catheters. \\
 & \feature{12355} & 0.87 & 4912 & Focus on endotracheal tube position and monitoring devices. \\
 & \feature{8757} & 0.92 & 420 & Central line or catheter placement described with specific position (e.g. 'mid SVC'). \\
 & \feature{14399} & 0.88 & 360 & Changes in Swan-Ganz catheter placement and position. \\
 & \feature{9328} & 0.91 & 2687 & Describes postsurgical changes after lung surgeries. \\
 & \feature{13592} & 0.96 & 416 & Presence and mention of surgical clips in the imaging reports. \\
 & \feature{12150} & 0.90 & 1040 & Presence of surgical clips in imaging unrelated to main findings. \\
 & \feature{14168} & 0.87 & 816 & Presence of suture material or surgical changes in lung fields. \\
 & \feature{13375} & 0.85 & 1188 & Presence of stents or vascular devices in chest imaging comparison. \\
 & \feature{14905} & 0.85 & 415 & Intact median sternotomy wires noted in findings. \\
 & \feature{3037} & 0.84 & 25498 & Descriptions of medical device positioning relative to anatomy, often compared to prior images. \\
 & \feature{5156} & 0.83 & 10791 & Evaluation of medical device placements and changes. \\
 & \feature{2475} & 0.83 & 11692 & Precise positioning or placement of medical devices in radiological reports. \\
 & \feature{65} & 0.83 & 10219 & Focus on placement and position of tubes and lines in radiological findings. \\
 & \feature{3246} & 0.80 & 27373 & Evaluation of pacemaker or ICD lead positions in chest X-rays. \\
 & \feature{5791} & 0.75 & 13374 & Analysis and repositioning of Dobbhoff tube placement. \\
 & \feature{1729} & 0.80 & 11306 & Assessment and positioning of endotracheal tubes. \\
\midrule
\multirow{2}{=}{Abnormal findings} & \feature{13963} & 0.94 & 258 & Attributes suggestive of chronic obstructive pulmonary disease (COPD). \\
 & \feature{12585} & 0.93 & 322 & Unfolded or tortuous thoracic aorta in radiology reports. \\
 & \feature{13113} & 0.87 & 478 & Unfolding or tortuosity of the thoracic aorta. \\
 & \feature{1336} & 0.86 & 11796 & Aortic tortuosity or calcification identified in chest imaging. \\
 & \feature{14586} & 0.94 & 533 & Presence of plate-like or linear atelectasis. \\
 & \feature{1374} & 0.88 & 3349 & Presence of atelectasis in imaging findings. \\
 & \feature{6105} & 0.88 & 3109 & Mentions and descriptions of atelectasis. \\
 & \feature{14427} & 0.86 & 443 & Retrocardiac opacification or atelectasis indicating volume loss or infection. \\
 & \feature{11555} & 0.93 & 401 & Detection of calcified structures in radiological findings. \\
 & \feature{9958} & 0.89 & 602 & Atherosclerotic calcifications noted at the aortic knob or thoracic aorta. \\
 & \feature{12236} & 0.89 & 656 & Presence of aortic arch calcifications or related cardiac calcifications. \\
 & \feature{4875} & 0.86 & 4444 & Cardiomegaly or enlarged cardiac silhouette. \\
 & \feature{6108} & 0.86 & 11089 & Findings of pulmonary vascular congestion or pulmonary edema. \\
 & \feature{13199} & 0.92 & 302 & Emphysema-related findings or descriptors. \\
 & \feature{11509} & 0.89 & 1123 & Observations of rib fractures in chest imaging reports. \\
 & \feature{13400} & 0.89 & 630 & Presence of pleural scarring or thickening, often unchanged, in radiology reports. \\
 & \feature{10636} & 0.86 & 743 & Blunting of costophrenic angles suggesting pleural effusion. \\
 & \feature{6412} & 0.84 & 5922 & Detection of pleural effusions on imaging studies. \\
 & \feature{13506} & 0.90 & 486 & Mention of scoliosis in radiological comparisons. \\
 & \feature{13515} & 0.97 & 159 & Elevation of the hemidiaphragm. \\
 & \feature{13911} & 0.90 & 412 & Elevation of the hemidiaphragm. \\
 & \feature{7458} & 0.82 & 14739 & Elevation of the hemidiaphragm on imaging studies. \\
 & \feature{13258} & 0.86 & 3238 & Presence of hiatal hernia and related structural effects. \\
 & \feature{1172} & 0.84 & 19306 & Cardiomegaly or heart enlargement observations in chest imaging. \\
 & \feature{6987} & 0.78 & 55984 & Cardiopulmonary findings suggesting congestion, cardiomegaly, or pleural effusion. \\
 & \feature{6343} & 0.80 & 40032 & Low lung volumes increasing broncho-vascular markings or heart silhouette. \\
 & \feature{2883} & 0.76 & 45007 & Changes or stability in pneumothorax or pneumomediastinum over time compared to prior studies. \\
\midrule
\multirow{2}{=}{Normal findings} & \feature{11891} & 0.87 & 1374 & Normal imaging findings with emphasis on absence of acute pathology. \\
 & \feature{10709} & 0.87 & 1761 & Findings indicate clear lungs with no signs of pleural effusion, pneumothorax, consolidation, or pulmonary edema. \\
 & \feature{8735} & 0.85 & 949 & Clear lungs, no pleural effusion or pneumothorax, unremarkable cardiac and mediastinal silhouettes. \\
 & \feature{10234} & 0.85 & 1347 & Clear lungs with no focal consolidation, effusion, or pneumothorax. \\
\midrule
\multirow{2}{=}{Temporal changes} & \feature{9876} & 0.86 & 800 & Comparison with prior images and mention of healed rib fractures. \\
 & \feature{1646} & 0.79 & 30532 & Interval change in disease findings from prior imaging. \\
 & \feature{1599} & 0.79 & 98759 & Describing findings without comparison to prior images. \\
 & \feature{6150} & 0.75 & 135961 & Reports describe interval changes in medical devices or effusions compared to prior images. \\
\midrule
\multirow{2}{=}{Textual features} & \feature{6907} & 0.91 & 5185 & Presence of double-bracketed image tags [[$<$image$>$]]. \\
 & \feature{12106} & 0.88 & 660 & Use of 'however' in clinical findings indicating possible issues needing further investigation. \\
 & \feature{7433} & 0.88 & 5661 & Prominent images marked by double brackets [[$<$image$>$]] in descriptions. \\
 & \feature{9473} & 0.86 & 677 & Use of 'possible' or 'possibly' indicating uncertainty. \\
 & \feature{10643} & 0.86 & 999 & Immediate notification of findings by telephone upon discovery. \\
 & \feature{11088} & 0.76 & 719 & No prior images available ('N/A') for comparison in the context. \\
 & \feature{3716} & 0.72 & 2466 & Presence of 'shortness of breath' as an indication. \\
 & \feature{9272} & 0.72 & 883 & Use of 'congestion' in radiology findings. \\
\bottomrule
\end{longtable}
\endgroup

\begin{table*}[t]
\caption{Additional examples of steered reports including evaluation. On-target changes between the original and the steered report are highlighted in \textcolor{ForestGreen}{green}, off-target effects in \textcolor{Orange}{orange}. The subtitle colours match the corresponding bars in \Cref{fig:pos-steering-all}}
\label{tab:more-steering-examples}
\footnotesize
\begin{tabular}{@{}p{0.13\linewidth} p{0.84\linewidth}@{}}
\toprule
\multicolumn{2}{c}{\textcolor{ForestGreen}{\bf Only on-target changes}}  \\
\midrule
\textbf{Feature} & \feature{1646}, $\fone=0.79$: \featexpl{Interval change in disease findings from prior imaging.} \\
\cmidrule{2-2}
\textbf{Original} & The heart size is mildly enlarged. The mediastinal contours are unremarkable. There is mild pulmonary vascular congestion. Small bilateral pleural effusions are noted. No pneumothorax is identified. There are no acute osseous
abnormalities. \\
\cmidrule{2-2}
\textbf{Steered} $\balpha~\mathbf{=10}$ & The heart size is mildly enlarged. The mediastinal contours are unremarkable. There is mild pulmonary vascular congestion, \textcolor{ForestGreen}{slightly worse when compared to the prior study}. Additionally, small bilateral pleural effusions are \textcolor{ForestGreen}{new compared to the prior exam}. No focal consolidation or pneumothorax is identified. There are no acute osseous
abnormalities. \\
\cmidrule{2-2}
\textbf{Scores} & on-target score=1.0 off-target score=0.0 \\
\midrule
\textbf{Feature} & \feature{10709}, $\fone=0.87$: \featexpl{Findings indicate clear lungs with no signs of pleural effusion, pneumothorax, consolidation, or pulmonary edema.} \\
\cmidrule{2-2}
\textbf{Original} & \textcolor{ForestGreen}{There is a new opacity in the right mid lung, which is concerning for pneumonia.} The cardiac, hilar and mediastinal contours are normal. No pleural abnormality is seen. \\
\cmidrule{2-2}
\textbf{Steered} $\balpha~\mathbf{=10}$ & \textcolor{ForestGreen}{The lungs are clear.} The cardiac, hilar and mediastinal contours are normal. No pleural abnormality is seen. \\
\cmidrule{2-2}
\textbf{Scores} & on-target score=1.0 off-target score=0.0 \\
\midrule[\heavyrulewidth]
\multicolumn{2}{c}{\textcolor{RoyalBlue}{\bf Both on- and off-target changes}}  \\
\midrule
\textbf{Feature} & \feature{6105}, $\fone=0.88$: \featexpl{Mentions and descriptions of atelectasis.} \\
\cmidrule{2-2}
\textbf{Original} & AP and lateral views of the chest. The lungs are hyperinflated but clear of consolidation. There is no effusion. Cardiomediastinal silhouette is within normal limits. No acute osseous abnormality detected. \\
\cmidrule{2-2}
\textbf{Steered} $\balpha~\mathbf{=10}$ & \textcolor{Orange}{PA} and lateral views of the chest were provided. The lungs are hyperinflated, though there is no focal consolidation, effusion, \textcolor{Orange}{or pneumothorax}. \textcolor{ForestGreen}{There is a linear density in the right lower lung which is most compatible with atelectasis}. Cardiomediastinal silhouette is \textcolor{Orange}{stable}. Bony structures are intact. \\
\cmidrule{2-2}
\textbf{Scores} & on-target score=1.0 off-target score=0.2 \\
\midrule
\textbf{Feature} & \feature{3246}, $\fone=0.80$: \featexpl{Evaluation of pacemaker or ICD lead positions in chest X-rays.} \\
\cmidrule{2-2}
\textbf{Original} & \textcolor{Orange}{A Port-A-Cath terminates in the superior vena cava}. A nasogastric tube \textcolor{Orange}{terminates in the stomach}. The cardiac, mediastinal and hilar contours appear stable. There is no pleural effusion or pneumothorax. The lungs appear clear. \\
\cmidrule{2-2}
\textbf{Steered} $\balpha~\mathbf{=10}$ & A nasogastric tube \textcolor{Orange}{terminates in the right atrium.} \textcolor{ForestGreen}{A dual-lead pacemaker/ICD device appears unchanged.} \textcolor{Orange}{The heart is normal in size.} The mediastinal and hilar contours appear stable. There is no pleural effusion or pneumothorax. The lungs appear clear. \\
\cmidrule{2-2}
\textbf{Scores} & on-target score=0.3 off-target score=0.7 \\
\midrule[\heavyrulewidth]
\multicolumn{2}{c}{\textcolor{Orange}{\bf Only off-target changes}}  \\
\midrule
\textbf{Feature} & \feature{13506}, $\fone=0.90$: \featexpl{Mention of scoliosis in radiological comparisons.}\\
\cmidrule{2-2}
\textbf{Original} & The cardiomediastinal and hilar contours are within normal limits. Lungs are well expanded and clear. There are no focal consolidations, pleural effusions, pulmonary edema or pneumothorax. \\
\cmidrule{2-2}
\textbf{Steered} $\balpha~\mathbf{=10}$ & The cardiomediastinal and hilar contours are within normal limits. \textcolor{Orange}{There is evidence of calcified mediastinal and hilar lymph nodes consistent with known sarcoidosis.} There is no focal consolidation, pleural effusion or pneumothorax. \textcolor{Orange}{Sternotomy wires are intact.} \\
\cmidrule{2-2}
\textbf{Scores} & on-target score=0.0 off-target score=0.5 \\
\midrule
\textbf{Feature} & \feature{14399}, $\fone=0.88$: \featexpl{Changes in Swan-Ganz catheter placement and position.}\\
\cmidrule{2-2}
\textbf{Original} & The heart size is within normal limits. The mediastinal and hilar contours are normal. The lungs are clear. There is no pleural effusion or pneumothorax. \\
\cmidrule{2-2}
\textbf{Steered} $\balpha~\mathbf{=10}$ & The heart size is within normal limits. The mediastinal and hilar contours are also within normal limits. \textcolor{Orange}{The lung volumes are low, but no consolidation}, pleural effusion or pneumothorax is seen. \\
\cmidrule{2-2}
\textbf{Scores} & on-target score=0.0 off-target score=0.2 \\
\bottomrule

\end{tabular}
\end{table*}

\begin{table*}[t]
\caption{Additional examples of steered reports including evaluation. On-target changes between the original and the steered report are highlighted in \textcolor{ForestGreen}{green}, off-target effects in \textcolor{Orange}{orange}. The subtitle colours match the corresponding bars in \Cref{fig:pos-steering-all}}
\label{tab:neg-steering-examples}
\footnotesize
\begin{tabular}{@{}p{0.14\linewidth} p{0.83\linewidth}@{}}
\toprule
\multicolumn{2}{c}{\textcolor{ForestGreen}{\bf Only on-target changes}}  \\
\midrule
\textbf{Feature} & \feature{6987}, $\fone=0.78$: \featexpl{Cardiopulmonary findings suggesting congestion, cardiomegaly, or pleural effusion.} \\
\cmidrule{2-2}
\textbf{Original} & As compared to the chest radiograph from 1 day prior, right-sided pigtail catheter has been removed. No pneumothorax. \textcolor{ForestGreen}{Small right-sided pleural effusion has slightly increased. Mild pulmonary vascular congestion. Mild cardiomegaly.} \\
\cmidrule{2-2}
\textbf{Steered} $\balpha~\mathbf{=-10}$ & As compared to the chest radiograph from 1 day prior, right-sided pigtail catheter has been removed. No pneumothorax.
The remaining appearance of the lungs is unchanged. \\
\cmidrule{2-2}
\textbf{Scores} & on-target score=1.0 off-target score=0.1 \\
\midrule
\textbf{Feature} & \feature{1646}, $\fone=0.79$: \featexpl{Interval change in disease findings from prior imaging.} \\
\cmidrule{2-2}
\textbf{Original} & The cardiac, mediastinal and hilar contours appear stable. There is no pleural effusion or pneumothorax. The lungs appear clear. Bony structures are unremarkable. \textcolor{ForestGreen}{There has been no significant change.} \\
\cmidrule{2-2}
\textbf{Steered} $\balpha~\mathbf{=-10}$ & The cardiac, mediastinal and hilar contours appear stable. There is no pleural effusion or pneumothorax. The lungs
appear clear. Bony structures are unremarkable. \\
\cmidrule{2-2}
\textbf{Scores} & on-target score=1.0 off-target score=0.0 \\
\midrule[\heavyrulewidth]
\multicolumn{2}{c}{\textcolor{RoyalBlue}{\bf Both on- and off-target changes}}  \\
\midrule
\textbf{Feature} & \feature{6105}, $\fone=0.88$: \featexpl{Mentions and descriptions of atelectasis.} \\
\cmidrule{2-2}
\textbf{Original} & \textcolor{Orange}{Right small pleural effusion has improved since \_} \textcolor{Orange}{with pigtail still in place}. There is no pneumothorax. \textcolor{ForestGreen}{Left lower lobe atelectasis has improved.} There is no new lung consolidation. Mediastinal and cardiac contours are normal. \\
\cmidrule{2-2}
\textbf{Steered} $\balpha~\mathbf{=-10}$ & \textcolor{Orange}{Right pigtail is no longer seen} and there is no visible pneumothorax. The lungs are clear. The cardiac and mediastinal contours are unremarkable. \textcolor{Orange}{There is no pleural effusion.} \\
\cmidrule{2-2}
\textbf{Scores} & on-target score=1.0 off-target score=0.5 \\
\midrule
\textbf{Feature} & \feature{2475}, $\fone=0.83$: \featexpl{Precise positioning or placement of medical devices in radiological reports.} \\
\cmidrule{2-2}
\textbf{Original} & As compared to the previous radiograph, there is no relevant change. \textcolor{ForestGreen}{The external pacemaker is in constant position.} There is no evidence of pneumothorax or other complication. No pleural effusions. No pneumonia. \textcolor{Orange}{Moderate cardiomegaly} without pulmonary edema.\\
\cmidrule{2-2}
\textbf{Steered} $\balpha~\mathbf{=-10}$ & As compared to the previous radiograph, there is no relevant change. \textcolor{Orange}{Left shoulder in situ.} No pneumothorax. No pulmonary edema. No pleural effusions. No pneumonia.\\
\cmidrule{2-2}
\textbf{Scores} & on-target score=1.0 off-target score=0.3 \\
\midrule[\heavyrulewidth]
\multicolumn{2}{c}{\textcolor{Orange}{\bf Only off-target changes}}  \\
\midrule
\textbf{Feature} & \feature{9995}, $\fone=0.85$: \featexpl{Pacemaker presence and description in chest imaging reports.}\\
\cmidrule{2-2}
\textbf{Original} & A left pectoral dual-lead cardiac pacemaker device is unchanged. The lungs are well expanded. No focal consolidation, edema, effusion, or pneumothorax. The heart is top-normal in size. The mediastinum is not widened. \textcolor{Orange}{Aortic knob calcifications are mild.} \\
\cmidrule{2-2}
\textbf{Steered} $\balpha~\mathbf{=-10}$ & A left pectoral dual-lead cardiac pacemaker device is unchanged. The lungs are well-expanded. No focal consolidation, edema, effusion, or pneumothorax. The heart is top-normal in size. The mediastinum is not widened. \textcolor{Orange}{The hila are unremarkable.}\\
\cmidrule{2-2}
\textbf{Scores} & on-target score=0.0 off-target score=0.2 \\
\midrule
\textbf{Feature} & \feature{12585}, $\fone=0.93$: \featexpl{Unfolded or tortuous thoracic aorta in radiology reports.}\\
\cmidrule{2-2}
\textbf{Original} & As compared to the previous radiograph, there is no relevant change. No evidence of pneumonia or other acute lung disease. \textcolor{Orange}{Minimal scars at the level of the right hilus. Known calcified granulomas in the right upper lobe.} Mild tortuosity of the thoracic aorta. No pleural effusions. No pneumothorax.\\
\cmidrule{2-2}
\textbf{Steered} $\balpha~\mathbf{=-10}$ & As compared to the previous radiograph, there is no relevant change. No evidence of pneumonia. \textcolor{Orange}{Borderline size of the
cardiac silhouette.} Mild tortuosity of the thoracic aorta. No pleural effusions. No pneumothorax. \\
\cmidrule{2-2}
\textbf{Scores} & on-target score=0.0 off-target score=0.4 \\
\bottomrule

\end{tabular}
\end{table*}

\end{document}